\newif\ifdraft
\draftfalse
\documentclass[10pt, conference, twocolumn]{IEEEtran}

\ifdraft
  \usepackage{drafts}
\else
  \usepackage[nodrafts]{drafts}
\fi

\usepackage{cite}
\usepackage{xspace}

\usepackage{hyperref}
\usepackage{subcaption}
\usepackage{tabularx}
\usepackage{figures}
\usepackage{graphicx,wrapfig,lipsum}
\usepackage{float}
\usepackage{paralist}
\usepackage{amsfonts}
\usepackage{amsmath}
\usepackage{multirow}
\usepackage[linguistics]{forest}
\usepackage{bm}
\usepackage[inline]{enumitem}
\usepackage[skip=0.15\baselineskip]{caption}

\newcommand{\man}{\textsc{Man}\xspace}
\newcommand{\gen}{\textsc{Gen}\xspace}

\pdfmapfile{=rtxr.map}

\begin{document}

\title{Automatic Generation of Atomic Consistency Preserving Search Operators for Search-Based Model Engineering}

\author{\IEEEauthorblockN{Alexandru Burdusel}
\IEEEauthorblockA{\textit{Department of Informatics} \\
\textit{King's College London} \\
30 Aldwych, London, WC2B 4BG \\
alexandru.burdusel@kcl.ac.uk}
\and
\IEEEauthorblockN{Steffen Zschaler}
\IEEEauthorblockA{\textit{Department of Informatics} \\
\textit{King's College London} \\
30 Aldwych, London, WC2B 4BG \\
szschaler@acm.org}
\and
\IEEEauthorblockN{Stefan John}
\IEEEauthorblockA{\textit{Department of Informatics} \\
\textit{Philipps-Universit\"{a}t Marburg} \\
Hans-Meerwein-Straße 6, Marburg, 35043 \\
stefan.john@uni-marburg.de}
}

\maketitle

\begin{abstract}


Recently there has been increased interest in combining the fields of Model\--Driven Engineering (MDE) and Search\--Based Software Engineering (SBSE). Such approaches use meta-heuristic search guided by search operators (model mutators and sometimes breeders) implemented as model transformations. The design of these operators can substantially impact the effectiveness and efficiency of the meta-heuristic search.
Currently, designing search operators is left to the person specifying the optimisation problem. However, developing consistent and efficient search-operator rules requires not only domain expertise but also in-depth knowledge about optimisation, which makes the use of model-based meta-heuristic search challenging and expensive.
In this paper, we propose a generalised approach to automatically generate atomic consistency preserving search operators (aCPSOs) for a given optimisation problem. This reduces the effort required to specify an optimisation problem and shields optimisation users from the complexity of implementing efficient meta-heuristic search mutation operators. We evaluate our approach with a set of case studies, and show that the automatically generated rules are comparable to, and in some cases better than, manually created rules at guiding evolutionary search towards near-optimal solutions.
This paper is an extended version of the paper with the same title published in the proceedings of the 22nd International Conference on Model Driven Engineering Languages and Systems (MODELS '19).

\end{abstract}

\begin{IEEEkeywords}
  model driven engineering, search based optimisation, search based model engineering, search based software engineering
\end{IEEEkeywords}

	\section{Introduction}
\label{section:introduction}

	
	Search-based software engineering (SBSE)~\cite{HarmanJones01} has seen increasing interest over the past decade. SBSE views software engineering as a problem of searching a, potentially very large, design space for optimal solutions and proposes techniques and tools for automating this search, typically using meta-heuristic search techniques. As a result, more design alternatives can be explored more quickly than would be possible manually. More recently, there has been an increasing interest in applying SBSE techniques in the context of MDE~\cite{Boussaid2017}, making the benefits of domain-specific modelling languages (DSMLs) available in an SBSE context. 
	
	Typical approaches (\emph{e.g.,}~\cite{Fleck2016SBMT,Hegedus+11}) use evolutionary algorithms. Users provide small endogenous model transformations (\emph{e.g.,} expressed as Henshin rules~\cite{HenshinDSL2017}) to specify mutation operators, which are then used for generating new candidate solution models.
	Writing these transformations is difficult: na\"ive implementations can easily cause the search to get stuck in local optima or to work very inefficiently.
	
	In this paper, we present a novel technique for automatically generating mutation operators from a declarative specification of an optimisation problem. 
	In particular, we generate operators that are consistency preserving, a key property for enabling the search to move out of local optima. We call such operators \emph{consistency preserving search operators} (CPSOs).
	
	We will show, through case-study--based experimental evaluation, that our automatically generated CPSOs result in search that is at least as efficient and effective as (and in some cases better than) search based on rules created manually. At the same time, automatic generation avoids the complexity and effort of manual creation and reduces the likelihood of erroneous or sub-optimal search operators being used. To the best of our knowledge, only~\cite{Struber2017} proposed an alternative approach for automatic generation of search operators, based on meta-learning. In contrast, our proposed technique avoids the need for a learning phase for each new problem.
		
	This paper generalises the work in~\cite{burdusel2017automatically}, where the authors explored initial ideas for rule generation in the context of a single case study without a generalised approach. Specifically, this paper makes the following contributions:
	\begin{enumerate}
		\item A general description and classification of CPSOs;
		\item An algorithm for generating atomic CPSOs (aCPSOs) that preserve multiplicity constraints; and
		\item An experimental evaluation with 3 case studies.
	\end{enumerate}	
	
	The remainder of this paper is structured as follows: In Sect.~\ref{section:background} we introduce some relevant background, followed by a running example in Sect.~\ref{section:running_example}. Section~\ref{section:proposed_method} contains the main contributions, describing CPSOs and the generation algorithm. Section~\ref{section:experiments} presents the experimental setup, followed by Sect.~\ref{section:results} in which we discuss results. In Sect.~\ref{section:related_work} we evaluate related work.

	\section{Background}
\label{section:background}


	In this section, we briefly describe relevant background to our research. In particular, we cover key MDE concepts, followed by an introduction to Search-Based Model Enginering (SBME) and a discussion of higher-order transformations. 

	\paragraph{Model-driven engineering}

		MDE considers models to be the primary artefact in software development~\cite{Bezivin2006}. Models are expressed in higher-level languages providing abstractions that are just right for the problem to be solved. Such languages are often called domain-specific modelling languages (DSMLs) and their (abstract) syntax is captured in metamodels (object-oriented models of the language concepts and their relationships). Model transformations---programs that take one or more models and produce new model(s) from them---are fundamental to MDE and to the powerful automation support it provides. Model transformations are often expressed using specialised languages and tools. Henshin~\cite{HenshinDSL2017} is one example, based on graph-transformation theory.

	\paragraph{Search-based model engineering}

		Search-based approaches in software engineering often use evolutionary search techniques. Evolutionary search (ES)~\cite{EibenEvolutionary2015} starts from a population of candidate solutions and \change{iterates these}{evolves these iteratively} by applying mutation (and possibly breeding) operators to generate new candidate solutions. In each \change{round}{evolution step}, all new candidate solutions' fitness is evaluated against the provided objective functions and this is used to rank solutions and select the best ones to carry over to the next \change{round}{generation}. This process is repeated until a given number of iterations is reached or a different stopping condition is met.
		A particular type of evolutionary algorithms are multi-objective evolutionary algorithms (MOEAs)~\cite{EibenEvolutionary2015}, which can handle multiple, possibly conflicting objective functions.
		A common problem with ES is that it may get stuck in so-called local optima; that is, solutions that are better than their neighbours (solutions that can be reached by a single application of a mutation operator) but that are not globally optimal.
		
		Evolutionary algorithms have been applied to MDE in two ways~\cite{John2019Searching, Boussaid2017}: some approaches (\emph{e.g.,}~\cite{Fleck2016SBMT,Hegedus+11}) encode candidate solutions as transformation chains and apply genetic algorithms to solve the search problems. Other approaches (\emph{e.g.,}~\cite{Zschaler16}) directly use models as candidate solutions. In both cases, model transformations are used to specify the available mutation operators. Fitness functions and constraints are specified as model queries using OCL or Java.

%
%

	\paragraph{Higher-order transformations}
		The term higher-order transformations (HOTs)~\cite{Tisi+09} refers to transformations that produce new model transformations. These are particularly useful when building advanced tools for MDE. In this paper, we are building on work on HOTs in two areas: generating consistency-preserving edit operations and generating model-repair transformations.

		In~\cite{Kehrer2016}, the authors introduce the SiDiff Edit Rule Generator (SERGe). SERGe is an Eclipse plugin to automatically generate consistency preserving edit operations (CPEOs), encoded as Henshin transformation rules, from an EMF metamodel. A CPEO is an atomic operation that, when applied to a consistent model instance, always generates a transformed consistent model instance. SERGe generates a complete set of CPEOs that can generate or delete \change{a}{any consistent} model instance through repeated applications. SERGe requires input metamodels to adhere to additional constraints on the supported multiplicities~\cite[Sect.~7.3.1]{KehrerPhD15}. Our rule-generation algorithm is based on the SERGe algorithm but additionally modifies the generated rules to ensure efficient search.

		The term model repair refers to the process of evolving an inconsistent model in order to make it consistent with its metamodel. In~\cite{Nassar2017}, the authors propose an approach for automatically generating repair operators encoded as Henshin rules, which can be used to repair an inconsistent model. The generated repair rules can be applied in a semi-interactive way to transform an invalid model into a valid instance of the metamodel. We make use of the catalogue of repair operations identified in~\cite{Nassar2017}.


    %

\paragraph{MDEOptimiser}

MDEOptimiser (MDEO) \footnote{\url{https://mde-optimiser.github.io}} is an SBME optimisation tool that allows users to specify optimisation problems in MDE using a DSL. The tool can be used as an Eclipse plugin as well as in standalone mode. The optimisation algorithms supported by the tool are implemented using MOEAFramework \footnote{\url{https://moeaframework.org}}.

The following elements describe an optimisation problem:

\begin{itemize}
	\item A problem metamodel describing the structure of problems and solutions;
	\item A set of solution constraints. These are either multiplicity constraints refining the problem metamodel multiplicities or additional well-formedness constraints implemented using OCL or Java;
	\item A set of endogenous model transformations typed over the problem metamodel, called mutation operators;
	\item A set of objective functions implemented as OCL or Java queries over solution models;
	\item A valid instance of the problem metamodel, providing initial problem constraints;
\end{itemize}

Based on these inputs, MDEO runs an ES. The input model is used as a seed for the initial population\delete{,} by making one copy for each population individual\delete{,} and then applying a random mutation to ensure variation. The tool uses the specified mutations to generate new candidate solutions in each algorithm step. Candidate solutions are evaluated after each generation, using the specified constraint and objective functions.

	\section{Running Example}
\label{section:running_example}

In this section, we introduce a running example of an SBME optimisation problem that can be specified using MDEO. Consider the scenario of a software development team who use Scrum as an agile software development methodology. Scrum, is a process management framework that proposes the use of fixed time iterations, also called sprints, during which a set of tasks defined as user stories are implemented, tested and released into the product under development \cite{rubin2012essential}.

We will briefly introduce the \change{key}{core} Scrum concepts as described in \cite{rubin2012essential}. The key artifacts of Scrum are the product, \add{the} product backlog and the sprint backlog. The product backlog is the list of all user stories that\add{,} when implemented\add{,} will result in a completed product. The sprint backlog is the list of user stories which the team aims to complete in a sprint. \change{Each user story has an estimated effort metric associated, also called story points, which denotes the required effort to complete the work.}{Each user story has associated story points, which serve as an estimate of the effort needed to complete it.}
%
%
The product owner is in charge of prioritising the backlog to make sure the most important user stories are worked on first. For the duration of a project, the development team completes several sprints. The average number of story points resulting from the completed user stories in a sprint is also known as team velocity.
 
In our example, we will consider that the user stories forming the backlog have an \emph{Importance} metric, denoting how important they are for a stakeholder, in addition to the \emph{Effort} metric, which shows the required effort for completion. The product owner is required to prioritise these tasks so that the average stakeholder importance is equally distributed across the sprints required to implement the work items in the backlog. We call this objective the Stakeholder Satisfaction Index and we calculate it as the standard deviation of average stakeholder importance across sprints.

\begin{figure}[!t]
\includegraphics[width=0.49\textwidth]{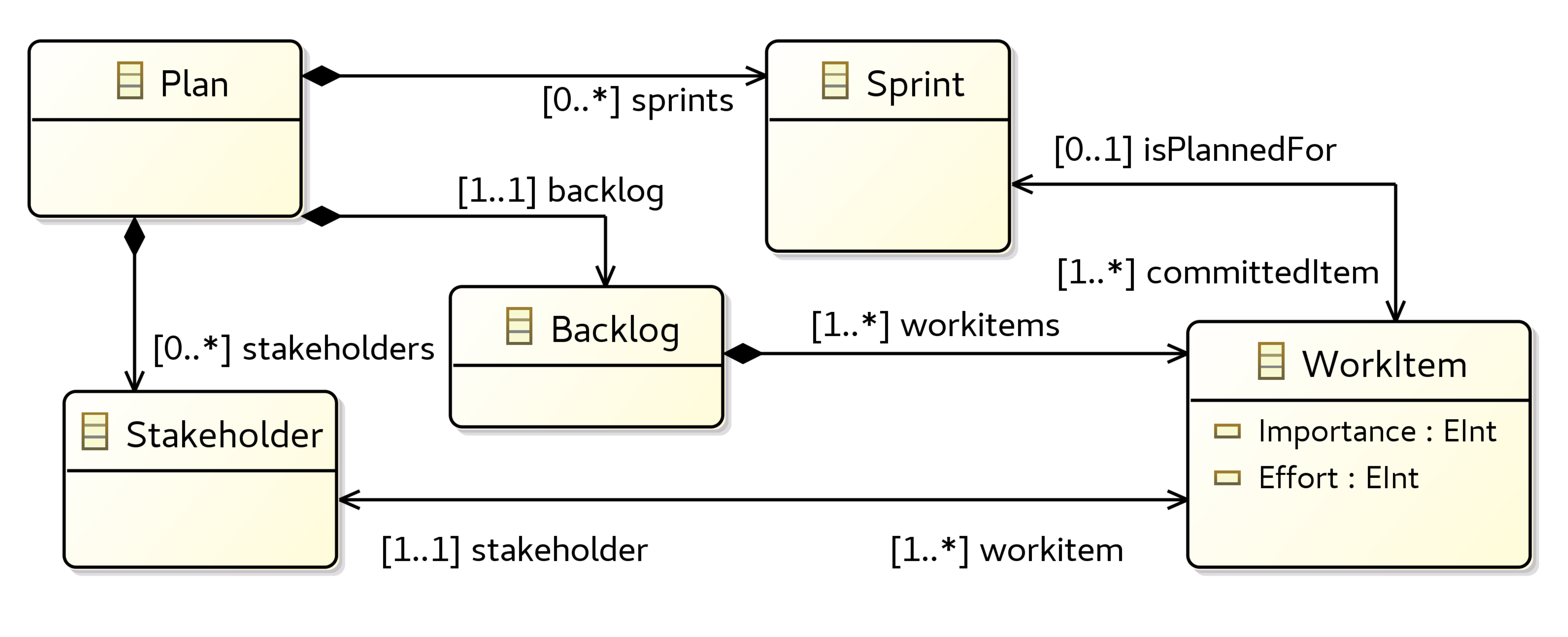}
\caption{Metamodel for the Scrum Planning problem}
\label{fig:sp-metamodel}
\end{figure}
	
In Fig.~\ref{fig:sp-metamodel} we show a metamodel of this problem. The goal of the problem is to assign \emph{WorkItem} elements to a number of \emph{Sprints} with the following objectives:
\begin{inparadesc}
	\item[Objective 1] minimise the \emph{Sprint} effort deviation;
	\item[Objective 2] minimise the Stakeholder Satisfaction Index.
\end{inparadesc}

The problem also has the following constraints: 
\begin{inparadesc}
	\item[Constraint 1] all \emph{WorkItem} entities must be assigned to a \emph{Sprint};
	\item[Constraint 2] no solution must have fewer \emph{Sprints} than total backlog \change{story points}{effort} divided by team velocity.
\end{inparadesc}


\begin{figure}[!b]
	\centering

	\subcaptionbox{Create Sprint\label{fig:create-sprint}}{\includegraphics[width=0.98\linewidth]{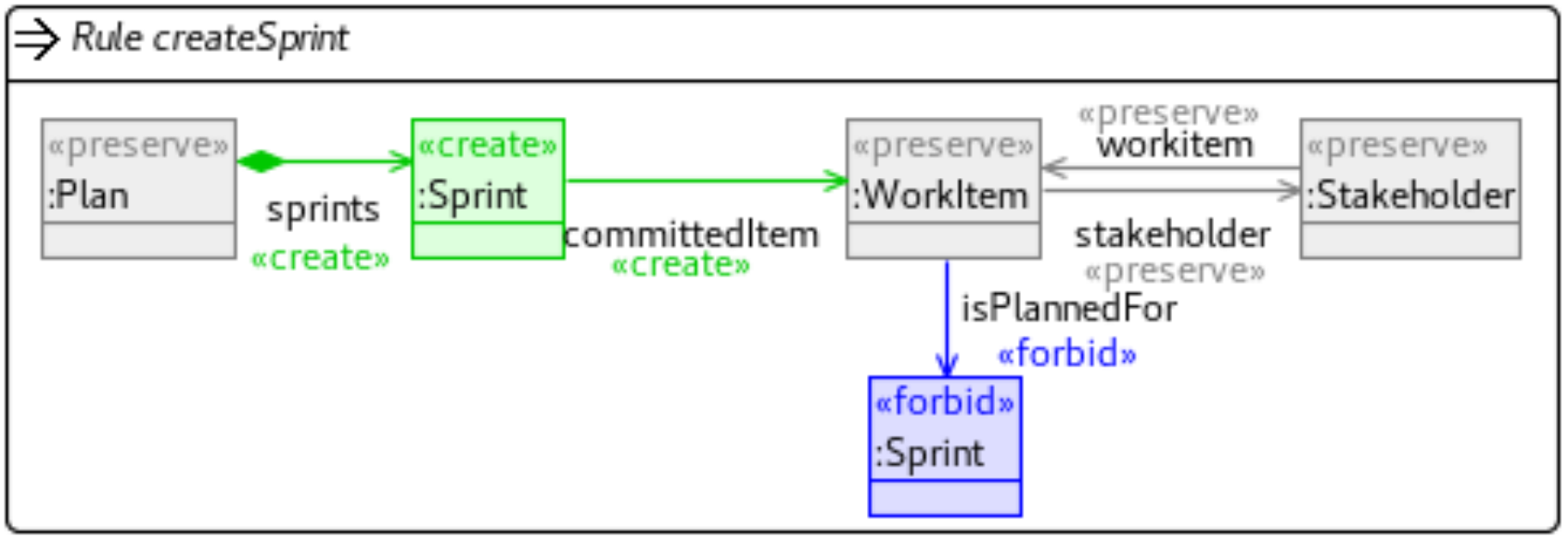}}%
	
	\subcaptionbox{Delete Sprint\label{fig:delete-sprint}}{\includegraphics[width=0.98\linewidth]{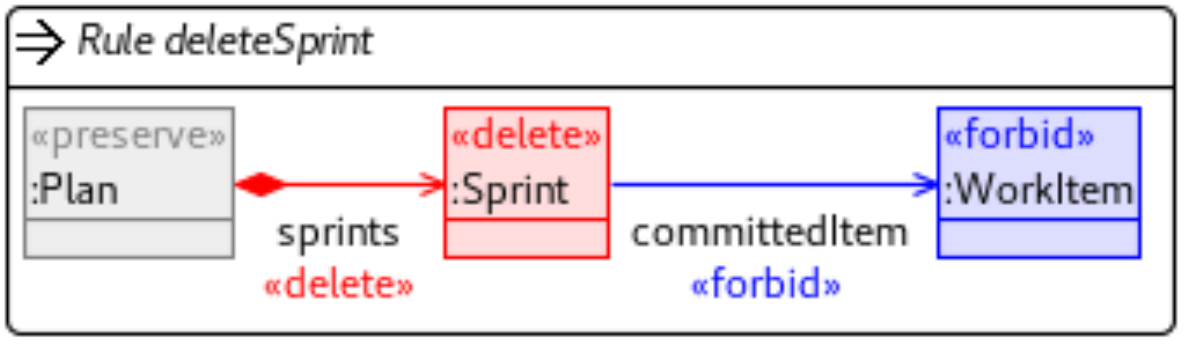}}%

	\subcaptionbox{Add WorkItem to Sprint\label{fig:add-item-to-sprint}}{\includegraphics[width=0.98\linewidth]{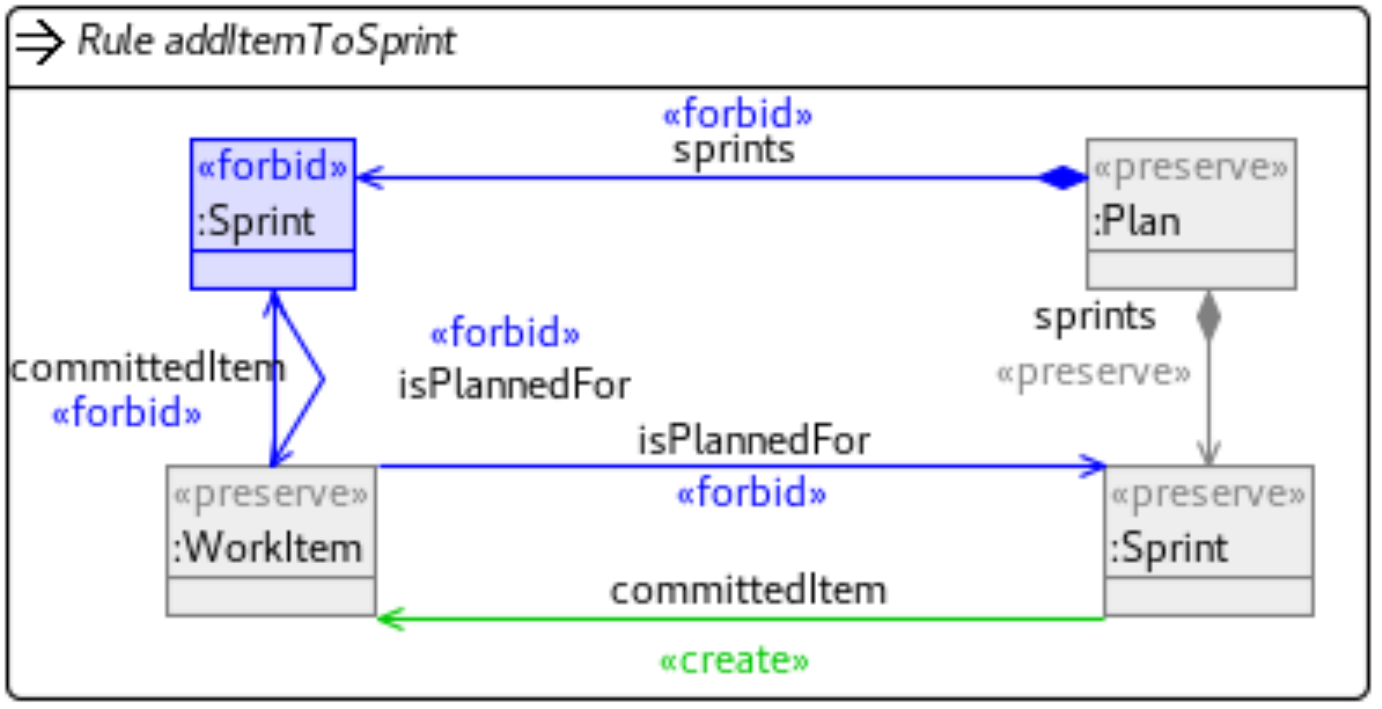}}%
	
	\subcaptionbox{Move WorkItem between Sprints\label{fig:move-work-item}}{\includegraphics[width=0.98\linewidth]{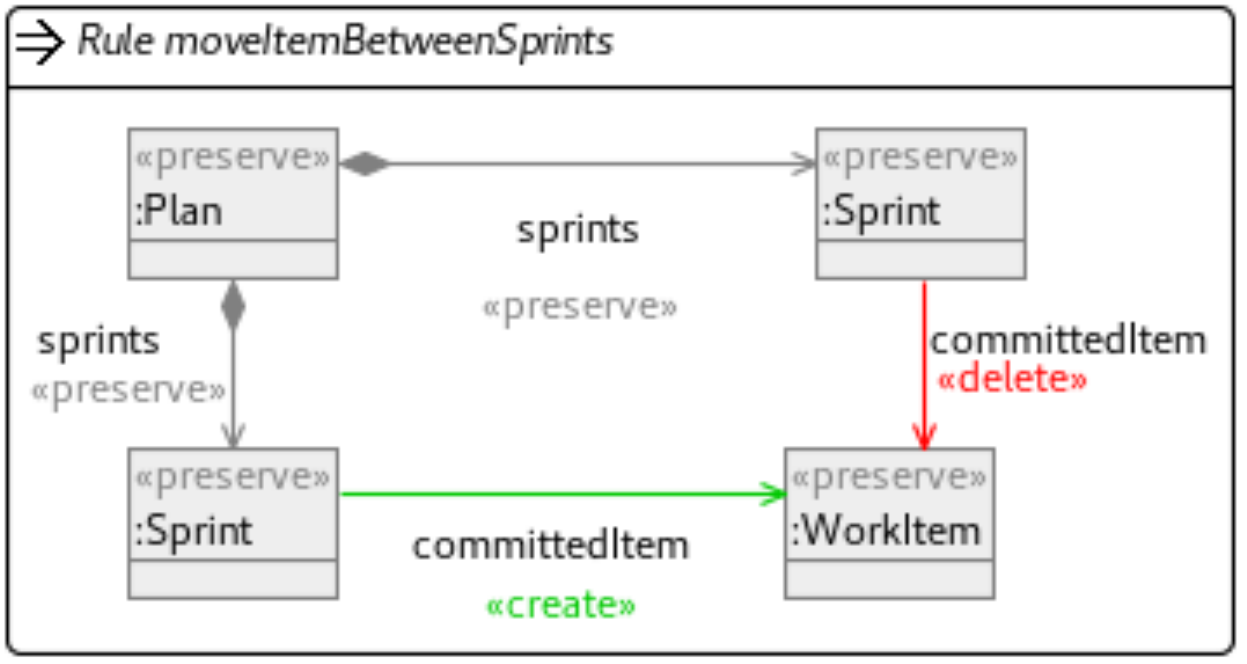}}%

	\caption{Summary of the mutation operators implemented manually for the Sprint Planning case study.}
	\label{fig:scrum-planning-manual-rules}
\end{figure}

To explore the search space of the Scrum Planning problem, the mutation operators must create \emph{Sprint} entities and assign \emph{WorkItem} elements to them, until all the \emph{WorkItem} elements belong to a \emph{Sprint}. \add{In Fig.~\ref{fig:scrum-planning-manual-rules} we include the mutation operators implemented manually for this case study.}

\delete{The structure of these operators is identical to the operators used by the authors in \cite{BurduselZschaler16}, because the changes permitted and the multiplicity constraints are identical between the two case studies.} \add{The solution constraints include a refined multiplicity (with a lower bound of 1) for the \emph{sprints} edge between a \emph{Plan} and \emph{Sprint} and also for the \emph{isPlannedFor} edge between a \emph{WorkItem} and a \emph{Sprint}.}

	\section{Generating Mutation Operators}
\label{section:proposed_method}


	Rather than asking the user to manually specify the mutation operators, our goal is to automatically generate them. In this section, we identify requirements for good mutation operators, introduce a general structure for mutation operators satisfying those requirements, and propose a systematic algorithm for generating them.

	As a result, a user will no longer be required to explicitly provide mutation operators as part of the optimisation problem specification. Instead, they will specify the sub\--metamodel for which mutation operators should be generated. This explicitly separates the parts of the metamodel that specify problem constraints from those which hold solution information. In our running example, the user would specify that the Sprint node and all its edges can be modified. This will produce rules that create new Sprints and assign \texttt{WorkItems} to them.

	\subsection{Requirements on mutation operators}
	\label{section:proposed_method:requirements}

		Generally, any transformation typed over the problem metamodel could be used as a mutation operator. However, here we are focusing on transformations that make small-granular changes (\emph{e.g.,} adding a node). This will allow a detailed exploration of the search space. 
		To identify additional requirements on mutation operators, we will explore two problems that can occur when operators are constructed na\"ively: getting stuck in local optima, and changing applicability of rules during different search phases.

		The search process can get stuck in \emph{local optima} when the constraints prevent the mutation operators from generating new and diverse individuals with a single transformation application.
		%
		Consider the Scrum planning use case including the following two operators: one for creating a new \texttt{Sprint} and one for moving a \texttt{WorkItem} from one \texttt{Sprint} to another.
		Once all the \texttt{WorkItem} elements have been assigned to a \texttt{Sprint}, no more new \texttt{Sprint} nodes can be created: because there are no more free \texttt{WorkItem} elements, the lower-bound constraint that no Sprint should be empty can no longer be satisfied for these new \texttt{Sprints}. If all the \texttt{WorkItems} have initially been assigned to a small number of \texttt{Sprints}, and no new \texttt{Sprints} can be created, the search will be unable to find solutions that have a good average distribution of \texttt{WorkItems} across the created \texttt{Sprints}.
		Note that creating two mutation operators, one to create an empty \texttt{Sprint} and one to move an existing \texttt{WorkItem} to the newly created \texttt{Sprint}, won't solve this problem: until the constraint is satisfied, the search algorithm would have to include the invalid solution in the archive and then apply the required repair operator in one of the following iterations. However, if all the other population individuals are valid, they will dominate the one with the invalid \texttt{Sprint}, which will be removed from the population.
		Generally, this problem is encountered where there are non-zero lower-bound multiplicities.
		In these cases, we require mutation operators to apply both edit and repair in one step.
		
		The search can be split into \emph{two phases:} in the first phase, all candidate solutions conform to the problem metamodel, but may not yet satisfy the additional solution constraints; in the second phase, all candidate solutions satisfy the additional solution constraints. These two phases potentially require different repair steps. Consider again a mutation operator creating a new Sprint node. In the first phase, the appropriate repair is to find a \texttt{WorkItem} that has not yet been assigned to another \texttt{Sprint} and assign it to the new \texttt{Sprint}. In the second phase, this rule is not applicable anymore, because no unassigned \texttt{WorkItems} remain. However, there is an alternative repair which takes a \texttt{WorkItem} from an existing Sprint with at least two \texttt{WorkItem} elements assigned to it. 
		We need to generate appropriate mutation operators for each phase of the search.
		
            
		Mutation operators that satisfy these requirements, we will call Consistency-Preserving Search Operators (CPSOs).

	\subsection{General structure of CPSOs}
	

		\begin{figure}
  			\begin{center}
			  \scalebox{0.85}{
				\begin{forest}
					[CPSO, fit=tight, s sep=0, l sep=1
					  [Edit Operation, align=center, l sep=1
					   [Atomic, align=center, l sep=1
						 [Node]
						 [Edge]
					   ]
					   [Compound]
					  ]
					  [\textbf{+}, no edge, before computing xy={s/.average={s}{siblings}}]
					  [Repair Operation, align=center, l sep=1
						[Atomic, align=center, l sep=1
						  [NAC]
						  [...]
						]
						[Iterative, before computing xy={s/.average={s}{siblings}}]
						[Recursive]
					  ]
					]
					\end{forest}
				}
			\end{center}
 			 \caption{CPSOs structure}
  			\label{fig:mutation-tree}
		\end{figure}
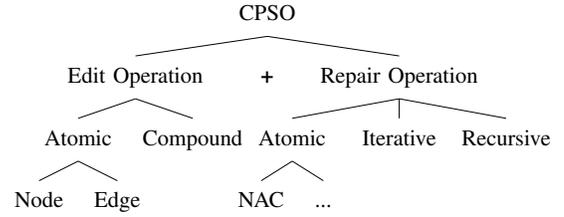
	
		As we have seen in the previous sub-section, CPSOs are transformation rules that combine a change to the model (an edit operation) with the necessary repair. In Fig.~\ref{fig:mutation-tree} we show the structure of CPSOs as well as further categorising edit and repair operations. We consider that edit operations can be either atomic or compound (a composition of multiple atomic operators). Atomic operators will either change a single node or a single edge. A repair operation can be atomic, iterative or recursive. Atomic repairs focus on a single edge and will not create or delete nodes beyond the original edit operation. An iterative repair is a combination of multiple atomic repairs for the same edit, for example where constraints on multiple edges would be broken by the edit. In contrast, a recursive repair creates or removes nodes as part of the repair, requiring recursive repair steps to be considered.
		In this paper we only consider atomic edit operations and atomic or iterative repair. We call the resulting operators atomic consistency preserving search operators (aCPSOs).
		

	
	\subsection{Generation algorithm}
	\label{section:proposed_method:generation_algorithm}



	In our current approach we focus only on multiplicity constraints. Supporting arbitrary constraints is not a trivial problem and it is beyond the scope of this paper to also support such constraints with our generation algorithm.

	In \cite{Kehrer2016,KehrerPhD15}, Kehrer et al.~introduce the concept of consistency preserving edit operations (CPEOs) and propose a mechanism for automatically generating them from a metamodel with multiplicity constraints. CPEOs can be used as CPSOs in cases where the solution metamodel only has open multiplicities. Any multiplicity is open if the lower bound is zero. Kehrer et al.'s mechanism does not support the generation of CPEOs for edges with closed multiplicities on both sides. Where only one multiplicity is closed, the mechanism only generates a limited range of repairs, which still causes the search to get stuck in local optima.

	\begin{figure}[!t]
		\centering
  \begin{center}
    \includegraphics[width=0.20\textwidth]{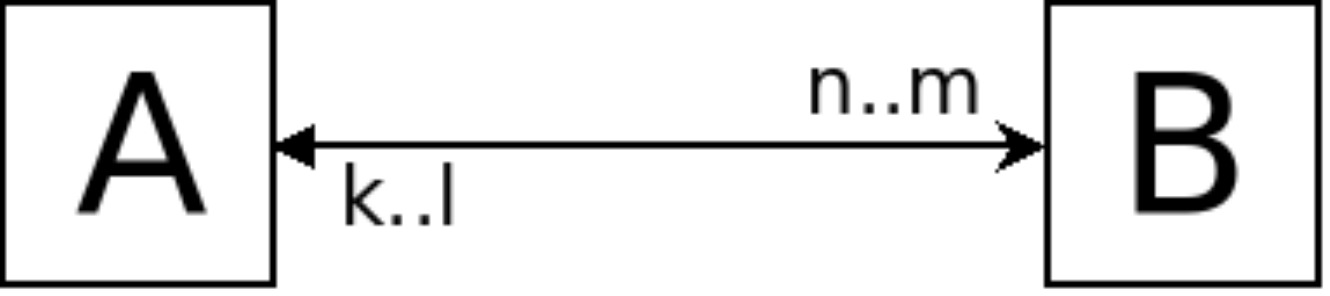}
  \end{center}
  \caption{Multiplicity patterns}
  \label{fig:multiplicity-patterns}
\end{figure}

		In this section we propose an algorithm to generate aCPSOs. We will structure the discussion based on the type of edit operations. For each edit operation we will then discuss relevant repair actions. We distinguish edit operations for nodes---namely create and delete---and for edges---add, remove, change, and swap. The available repair operations depend on the multiplicity pattern.
		Fig.~\ref{fig:multiplicity-patterns} shows the labels we will use in our discussion below.
		
		For each multiplicity pattern we consider, we aim to generate the minimal set of rules that would allow the search to avoid getting stuck in local optima. The minimal set of rules, ensures that we can perform all the create and delete node operations and add and remove edges between graph elements.




		\begin{figure}[!b]
			\centering

			\subcaptionbox{Create Node Rule\label{fig:create-node}}{\includegraphics[width=0.98\linewidth]{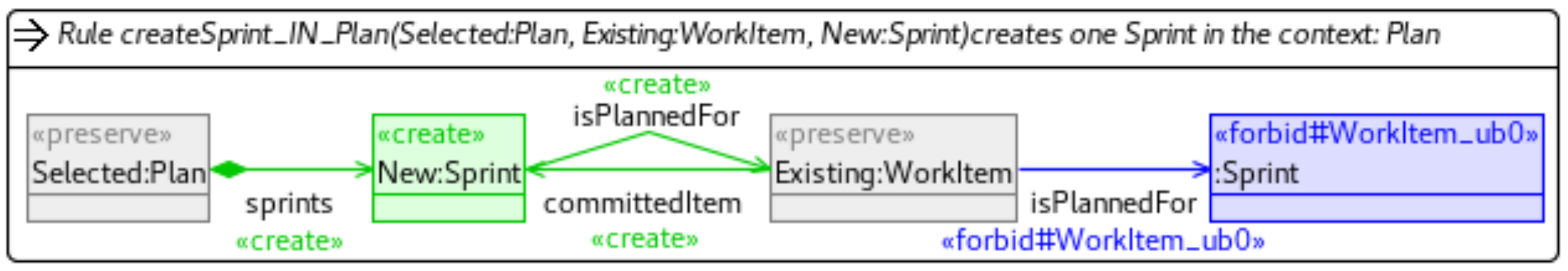}}%
			
			\subcaptionbox{Create Node LB Repair Rule\label{fig:create-node-lb-repair}}{\includegraphics[width=0.98\linewidth]{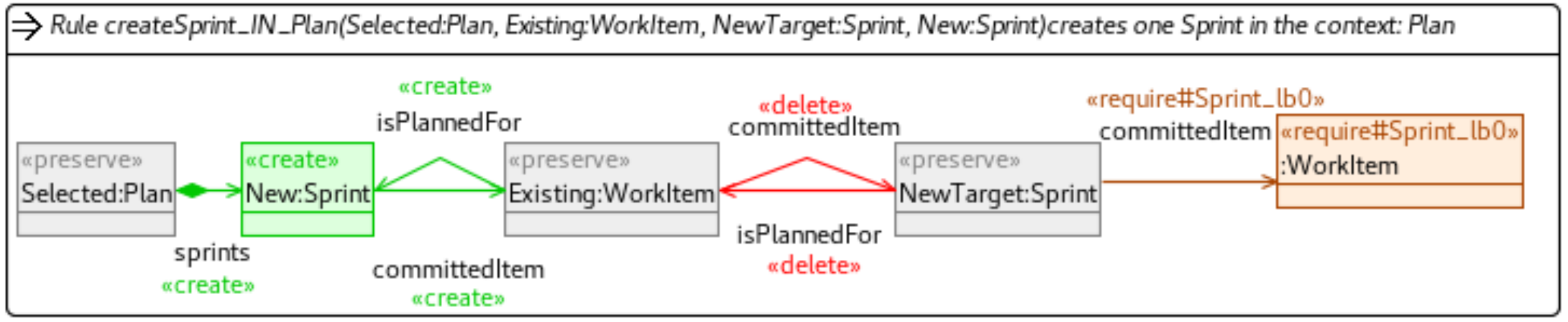}}%

			\subcaptionbox{Delete Node Rule\label{fig:delete-node}}{\includegraphics[width=0.98\linewidth]{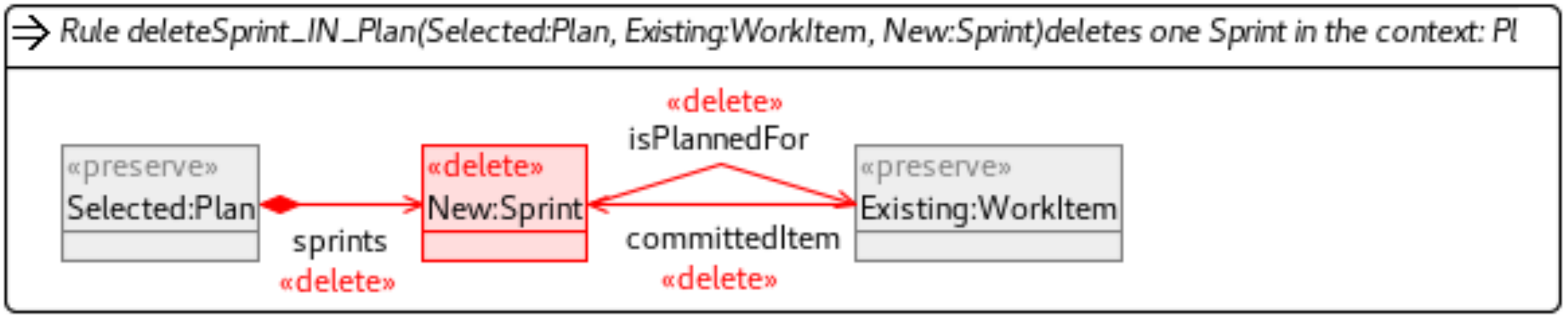}}%
			
			\subcaptionbox{Delete Node LB Repair Rule\label{fig:delete-node-lb-repair}}{\includegraphics[width=0.98\linewidth]{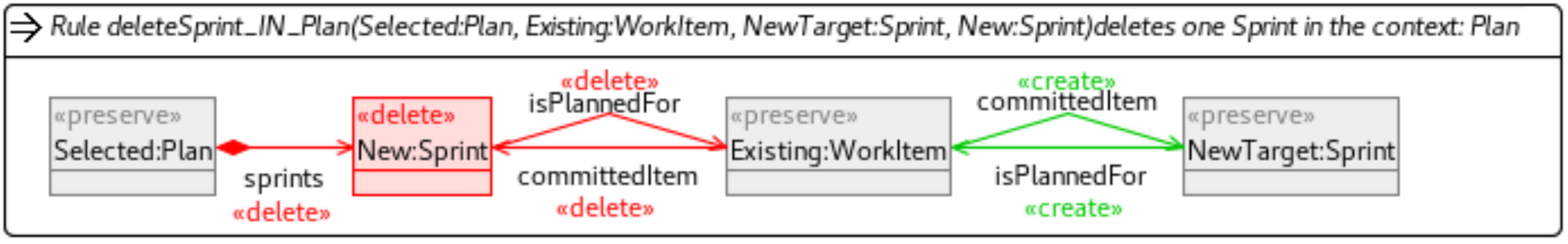}}%

			\caption{Generated node manipulation aCPSOs for the Scrum Planning case study encoded as Henshin model transformations.}
			\label{fig:scrum-planning-acpsos}
		\end{figure}

		\subsubsection{Manipulating nodes}
			
			In this section we describe the repair operations required for manipulating nodes. The types of aCPSOs that we generate for this, are composed of the atomic rule to create (delete) a node and a repair operation to connect (disconnect) the created (deleted) node to (from) mandatory neighbours (B nodes). The choice of repairs that can be applied is given by the multiplicity pattern between the node being edited and its neighbours. For some repairs there are many variants in-between, however we seek to minimise the number of generated rules, so we only generate the rules described. Kehrer et al. support a restricted set of multiplicities for creating and deleting a node of type A: (k,l) to (0,1) or (k, l) to (0,*). 

\begin{table}[!t]
\centering
\caption{Create node aCPSOs. In the table, `c' stands for `create', `lb r' for `lower-bound repair', and `f\#l' for `forbid l'.}
\label{tab:node-create-acpso}
\scalebox{0.8}{
\begin{tabular}{|l|l|l|l|l|}
\hline
                                                                                               & \textbf{n=0}                  & \textbf{n = 1 and m \textgreater n} & \textbf{n \textgreater 1 and m \textgreater n} & \textbf{n = m} \\
\hline
\textbf{\begin{tabular}[c]{@{}l@{}}k $\geq$ 0\\ l \textgreater k\\ l \textless *\end{tabular}} & \multirow{3}{*}{c A}                              & \begin{tabular}[c]{@{}l@{}}c A add n B (f\#l A)\\ c A lb r single B\end{tabular} & \begin{tabular}[c]{@{}l@{}}c A add n B (f\#l A)\\ c A lb r single B\\ c A lb r many B\end{tabular} & c A add n B (f\#l A) \\
\cline{1-1}\cline{3-5}
\textbf{\begin{tabular}[c]{@{}l@{}}k $\geq$ 0\\  l = *\end{tabular}}                              &           & \multicolumn{3}{c|}{c A add n B}\\ 
\cline{1-1}\cline{3-5}
\textbf{k = l}                                                                                          &           & c A lb r single B                                                                & \begin{tabular}[c]{@{}l@{}}c A add n B\\ c A lb r single B\\ c A lb r many B\end{tabular}          & N/A                  \\ \hline
\end{tabular}
}
\end{table}
			 
			\paragraph{Creating a node}

			In this section we introduce the types of repair operations generated for creating a node. For each repair we include the multiplicity patterns for which the generated repair is applicable. We include a summary of the generated rules in Table~\ref{tab:node-create-acpso}.
			            
            \begin{itemize}
            	\item \emph{NAC repair}: The first type of aCPSO that we generate, is for creating nodes that have a multiplicity pattern with ($n > 0$). 
				For this case, we generate a rule to create a node of type A and connect it to \emph{n} existing nodes of type B. If $l < *$, then a negative application condition (NAC) is added for the connected nodes B, to ensure that no upper-bound multiplicity invalidations occur (no more than \emph{l} nodes of type A assigned for each B). Nodes that have an open multiplicity don't need a repair operation.
				
				\add{Fig.~\ref{fig:create-node} shows an example of this aCPSO, generated for creating a \emph{Sprint} node, that is connected to a note of type \emph{WorkItem}. The rule includes a NAC for the \emph{WorkItem} node which requires that the \emph{WorkItem} node is not already assigned to a \emph{Sprint} node.}
                
                \item \emph{Single source lower bound repair}: The second type of aCPSO for creating a node, is for creating nodes that have a multiplicity pattern with ($n > 0$) and ($l < *$). This pattern means that A must have at least \emph{n} nodes of type B assigned to it and node B can have a limited number of nodes of type A assigned to it.
    		    We generate a rule to create a node of type A, and connect it to \emph{n} nodes of type B. Then, we generate a repair to satisfy the lower-bound for the created node A and repair the upper-bound for the existing n nodes of type B, by deleting the edges between the required \emph{n} nodes of type B from a \emph{single} existing node of type A and creating edges between them and the newly created node of type A. A positive application condition (PAC) is generated for the existing node A to ensure that the lower-bound multiplicity is not broken by this operation.

				\add{In Fig.~\ref{fig:create-node-lb-repair} we show an example of this aCPSO, generated for creating a \emph{Sprint} node, when all \emph{WorkItems} are already assigned to other \emph{Sprints}. The rule includes a PAC for the existing \emph{Sprint} node from which the \emph{WorkItem} node used for the repair is taken, to make sure that the lower-bound multiplicity is not invalidated.}

           		\item \emph{Multiple sources lower bound repair}: The third type of aCPSO for creating a node, is for creating nodes that have a multiplicity pattern with ($n > 1$) and ($l < *$). This pattern means that A must have at least \emph{n} nodes of type B assigned to it and node B can have a limited number of nodes of type A assigned to it.
           		For this case, we generate a rule to create a node of type A, and connect it to \emph{n} nodes of type B. Then, we generate a repair to satisfy the lower-bound for the created node A and repair the upper-bound for the existing n nodes of type B, by deleting the edges between the required \emph{n} nodes of type B from \emph{n} existing nodes of type A, and creating edges between them and the newly created node of type A. A PAC is generated for the existing nodes of type A to ensure that the lower-bound multiplicity is not broken by this operation. 
           		          
           \end{itemize}
			
			
			For node pairs that have a fixed multiplicity ($ n = m \wedge k = l $), at both ends of any edge, we do not generate a create node aCPSO. Any repair operation for this case requires the creation of the nodes at the opposite end of the edge, and thus a recursive repair.

\begin{table}[!t]
\centering
\caption{Delete Node aCPSOs. In the table, `d' stands for `delete', `r lb sg' for `repair lower bound single', `r lb mn' for `repair lower bound multiple', and `f\#m' for `forbid m'.}
\label{tab:node-delete-acpso}
\scalebox{0.8}{
\begin{tabular}{|l|l|l|l|}
\hline
                                                                                     & \textbf{m \textgreater\ n and m \textless\ *} & \textbf{m = *}                                                                                              & \textbf{n = m} \\ \hline
\textbf{k = 0}                                                                       & \multicolumn{3}{c|}{d A}                                                                                                                                                                                            \\ \hline
\textbf{\begin{tabular}[c]{@{}l@{}}k \textgreater 0\\ l \textgreater k\end{tabular}} & \multicolumn{3}{c|}{d A (require each B still has \#k A)}                                                                                                                                                                      \\ \hline
\textbf{k=l=1}                                                                       & d A r lb sg B (f\#m A) & d A r lb sg B                                                         & N/A            \\ \hline
\textbf{k=l \textgreater 1}                                                                         & \begin{tabular}[c]{@{}l@{}}d A r lb sg B (f\#m A)\\ d A r lb mn B (f\#m A)\end{tabular} & \begin{tabular}[c]{@{}l@{}}d A r lb sg B\\ d A r lb mn B\end{tabular} & N/A            \\ \hline
\end{tabular}
}
\end{table}

			\paragraph{Deleting a node}
			As with the description for the create operations, we divide the explanation based on repair type. We include a summary of the generated rules in Table~\ref{tab:node-delete-acpso}.
			
			\begin{itemize}
			
            	\item \emph{PAC repair}: The first type of aCPSO that we generate, is for deleting nodes that have a closed multiplicity ($k > 0$). This pattern means that B must have at least \emph{k} nodes of type A assigned and each node of type A must be assigned to at least  \emph{n} nodes of type A. 
				For this case we generate a rule to delete a node of type A and for each of its connected nodes of type B, a PAC is added to ensure that no lower-bound multiplicity invalidations occur after the deletion of the A node. This rule is not generated for cases where ($k=l$).
				
				\add{In Fig.~\ref{fig:delete-node} we include an example of this aCPSO, generated for deleting a \emph{Sprint} node, that has a \emph{WorkItem} assigned to it. For this example rule, there is no PAC generated for the \emph{WorkItem} because there is no lower-bound multiplicity limit.}
                
                \item \emph{Single target lower bound repair}: This type of aCPSO for deleting a node, is for deleting nodes that have a multiplicity pattern with ($k = 1$). This pattern means that each node of type B must be assigned to at least  \emph{k} nodes of type A.
                For this case, we generate a repair to satisfy the lower-bound for the k nodes B, 
				%
				%
				by creating edges between them and another single existing node of type A. A NAC is generated for the existing node A to ensure that the upper-bound multiplicity is not broken if ($m < *$).
				
				\add{Fig.~\ref{fig:delete-node-lb-repair} shows an example of this aCPSO, generated for deleting a \emph{Sprint} node, that has a \emph{WorkItem} assigned to it. For this example rule, there is no NAC generated because there is no upper-bound multiplicity limit.}
                
                \item \emph{Multiple target lower bound repair delete}: This type of aCPSO for deleting a node, is for deleting nodes that have a multiplicity pattern with ($k = l$) and ($l > 1$). This pattern means that A must have at least \emph{n} nodes of type B assigned and each node of type B must be assigned to at least  \emph{k} nodes of type A.
                For this case, we generate a repair to satisfy the lower-bound for node B, by creating edges between them and another existing \emph{n} nodes of type A. If required, a NAC is generated for the existing nodes of type A to ensure that the upper-bound multiplicity is not broken if ($m < *$). We only generate this rule for the case where exactly \emph{n} nodes of type B are attached to the A node to be deleted.
                
            \end{itemize}
			
			For node pairs that have a fixed multiplicity ($n = m \wedge k = l$), at both ends of any edge, we do not generate a delete node aCPSO. Similar to the create node operations, a repair operation for this case requires the deletion of the node at the opposite end of the node being deleted. We regard this type of operation as recursive, which we will look at in future work.
	
			\begin{figure}[!b]
				\centering
				\subcaptionbox{Add Edge Rule\label{fig:add-edge}}{\includegraphics[width=0.98\linewidth]{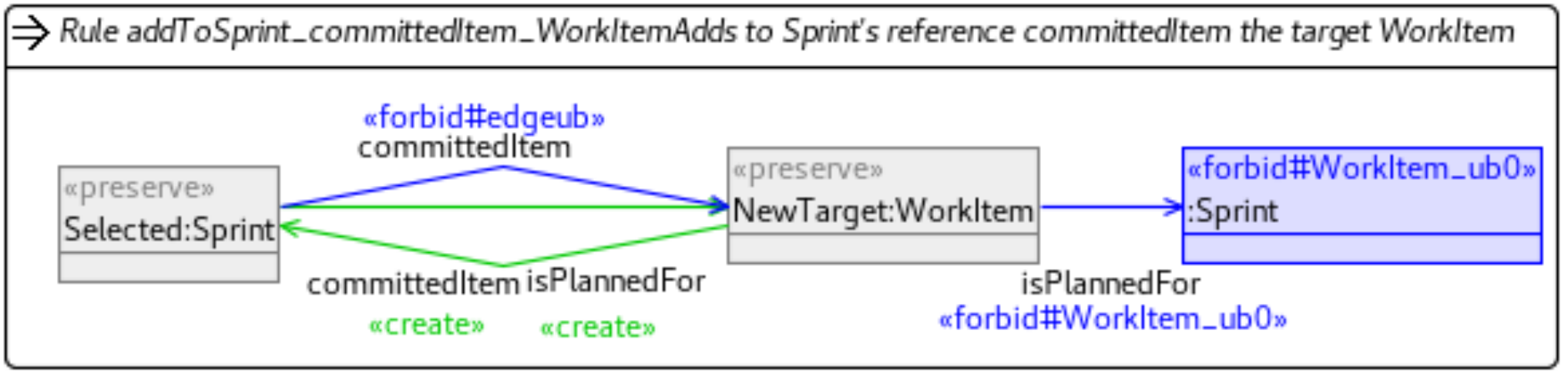}}%
	
				\subcaptionbox{Remove Edge Rule\label{fig:remove-edge}}{\includegraphics[width=0.98\linewidth]{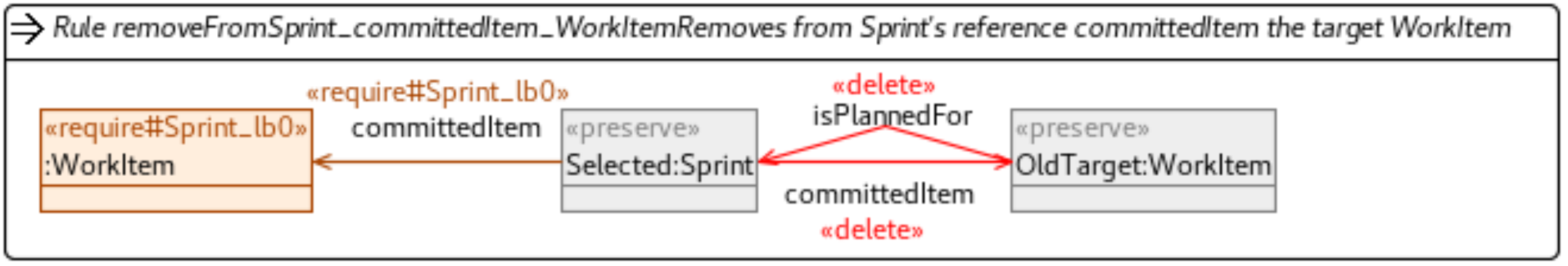}}%
	
				\subcaptionbox{Change Edge Rule\label{fig:change-edge}}{\includegraphics[width=0.98\linewidth]{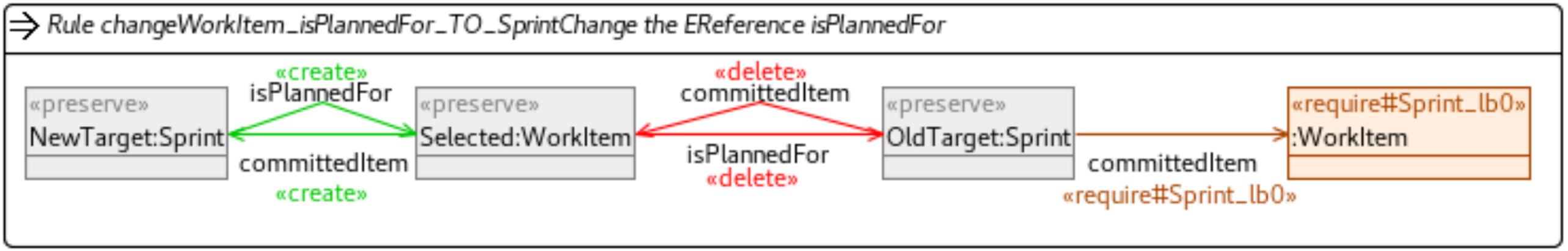}}%
				\caption{Generated edge manipulation aCPSOs for the Scrum Planning case study encoded as Henshin model transformations.}
				\label{fig:scrum-planning-acpsos}
			\end{figure}

		\subsubsection{Manipulating edges}

			In this section we show the types of aCPSOs we generate for manipulating edges between two nodes. Namely to add and to remove an edge from a node, together with corresponding repair operations. The add and remove edge operations are composed to obtain the more complex change and edge-swap operations. A complete list of the generated edge aCPSOs is included in Tables~\ref{tab:add-edge-acpso} and~\ref{tab:remove-edge-acpso}.

\begin{table}[!t]
\centering
\caption{Add-edge aCPSOs. In the table (P/N) denotes the presence of optional PAC and NACs that may be required by the source or target node multiplicity.}
\label{tab:add-edge-acpso}
\scalebox{0.9}{
\begin{tabular}{|l|l|l|l|}
\hline
                          & \textbf{m \textless *} & \textbf{m = *}            & \textbf{n=m}   \\ \hline
\textbf{l \textless *}    & Add edge NAC A B       & Add edge NAC B            & Swap edge      \\ \hline
\textbf{l = *}            & Add edge NAC A         & Add edge                  & Swap edge      \\ \hline
\textbf{k = l}            & Change edge (P/N A)            & Change edge  (P/N A)              & Swap edge     \\
\hline \end{tabular}
}
\end{table}

\begin{table}[!t]
\centering
\caption{Remove-edge aCPSOs. In the table (P/N) denotes the presence of optional PAC and NACs that may be required by the source or target node multiplicity.}
\label{tab:remove-edge-acpso}
\scalebox{0.9}{
\begin{tabular}{|l|l|l|l|}
\hline
                          & \textbf{n = 0}         & \textbf{n \textgreater 0} & \textbf{n = m} \\ \hline
\textbf{k = 0}            & Remove edge            & Remove edge PAC A         & Swap Edge      \\ \hline
\textbf{k \textgreater 0} & Remove edge PAC B      & Remove edge PAC AB        & Swap Edge      \\ \hline
\textbf{k = l}            & Change edge (P/N A)           & Change edge (P/N A)             & Swap Edge      \\ \hline
\end{tabular}
}
\end{table}

			\paragraph{Adding an edge}
			The aCPSO to add an edge between two existing nodes is identical to the add edge CPEO generated by Kehrer et al. This aCPSO includes a NAC to avoid invalidating any upper-bound constraints between the source and target nodes. This aCPSO is generated for all multiplicity patterns except for cases having a fixed multiplicity at one end or at both ends of the connected nodes ($~n = m \vee k = l$). 
			
			\add{In Fig.~\ref{fig:add-edge} we include an example of an aCPSO that adds an edge between a \emph{Sprint} and a \emph{WorkItem} with a NAC, fordiding that the two nodes are already connected.}
			
			\paragraph{Removing an edge}
			This aCPSO is also similar to a CPEO that Kehrer et al. generate, consisting of an operation to remove an edge between two existing nodes A and B. 

			The generated aCPSO includes a NAC to avoid invalidating any lower-bound constraints between the source and target nodes. This aCPSO is generated for all multiplicity patterns except for cases having a fixed multiplicity at one end or at both ends of the connected nodes ($~n = m \vee k = l$). 
			
			\add{In Fig.~\ref{fig:remove-edge} we include an example of an aCPSO that removes an edge between a \emph{Sprint} and a \emph{WorkItem} with a PAC, requiring that after the application of this rule, the \emph{Sprint} node still has at lease one \emph{WorkItem} node still assigned to it, to satisfy the lower-bound multiplicity.}
			
			\paragraph{Changing an edge}
			A change edge aCPSO moves a node of type B to another node of type A when node B has a fixed multiplicity.
			This type of operation, moves a node of type B with a lower bound multiplicity pattern ($k > 0$), to another node of type A, without invalidating the multiplicity constraints. The generated aCPSO includes PAC and NAC conditions to ensure that after the rule application, no lower-bound or upper-bound multiplicities are invalidated, for the source and target nodes respectively of type A (to ensure that no node has too many or to few nodes of type B after this rule application). 
			This aCPSO is generated for closed multiplicity patterns where a multiplicity pattern for either of the connected nodes is fixed (\emph{e.g.,}$~n = m \vee k = l$). 

			\add{Fig.~\ref{fig:change-edge} shows an example of this aCPSO, generated for changing an edge between a \emph{WorkItem} and two \emph{Sprints}. The rule includes a PAC for the \emph{Sprint} element from which the \emph{WorkItem} element is unassigned, to ensure that the lower-bound multiplicity of this node is not invalidated after the application of the rule.}
			
			\paragraph{Swapping two edges}
			An edge swap aCPSO is generated for fixed multiplicity patterns on the A side ($n = m$). This operation exchanges two nodes, between two pairs of similar node types. 
			For two existing, connected nodes A and B, the aCPSO, finds two other nodes of the same type, $A^\prime$ and $B^\prime$ and disconnects node A from node B and $A^\prime$ from $B^\prime$, and connects node A to $B^\prime$ and $A^\prime$ to B.
			
		\subsubsection{Iterative repair}
		Iterative repair rules are generated by creating combinations of the possible repair types described above, for all the edges of a node that has to be mutated. This approach increases the number of rules generated for nodes that have multiple edges. 
		
	\subsection{Running search with aCPSOs}
	\label{section:proposed_method:running_search}
	
		The algorithm proposed in the previous sub-section generates operators that preserve the consistency of the models modified. This addresses the first requirement on mutation operators that we identified in Sect.~\ref{section:proposed_method:requirements}. It does not yet address the second requirement, that mutation operators should work in both phases of an evolutionary search: phase 1, when some candidate solutions may not yet fully satisfy the solution-metamodel constraints, and phase 2, when all candidate solutions satisfy all solution-metamodel constraints. To satisfy this second requirement, we run the algorithm from Sect.~\ref{section:proposed_method:generation_algorithm} twice: First, we use it to generate rules based on problem-metamodel constraints. Next, we generate rules based on solution-metamodel constraints. We then use the union of the two sets of rules as the set of mutation operators for the evolutionary search.

	\section{Experiments}
\label{section:experiments}

	To evaluate our rule generation approach, we ran a number of experiments with a range of case studies. 
	The aim of these experiments is to show that the generated mutation operators are at least as good as a set of operators created manually. The automatic generation of transformations is already an improvement over the manual process, as we remove the error prone process of manual rule creation. \change{Our evaluation aims to show that there is no loss in search performance from generated operators.}{Our evaluation aims to investigate whether there is loss in search performance from generated operators.}
	For each case study, we prepared a set of manually created mutation operators. Then, we configured MDEOptimiser to automatically generate mutation operators. Using both pairs of mutation operators we ran experiments to solve the same problem instances, and we compare the results from the two approaches to validate that the solutions obtained using the automatically generated aCPSOs are comparable with the results obtained using manually implemented search operators.

	\change{In our experiments we are not including a direct comparison between our tool and other tools}{We are not including a comparison between our tool and other tools}. Such a comparison is beyond the scope of this paper. In \cite{John2019Searching} the authors compare the performance of MDEOptimiser and MoMOT, another model search tool that encodes search solutions as transformation chains.

\subsection{Case studies}
\label{subsection:case_studies}

	We have chosen a set of combinatorial optimisation problems that have been implemented using MDEOptimiser. In the following subsections, we include a brief description of each case study.

\subsubsection{Class-Responsibility Assignment}

	The Class Responsibility Assignment (CRA) case study has been introduced at the 2016 Transformation Tool Contest (TTC) \cite{Fleck+16_TTC_Case}. The goal of this problem is to transform a procedural software application to an object oriented architecture while maintaining good cohesion and coupling. The quality of the produced solutions is measured using the CRA index defined in~\cite{Fleck+16_TTC_Case}, as a single objective.
	The problem supplies a responsibility dependency graph, that contains a set of functions and attributes with dependencies between them. In the metamodel, these entities are instances of the abstract type Feature. 

	To solve this problem, the user is required to create \texttt{Classes} in the \texttt{ClassModel} and assign \texttt{Features} to them such that: all \texttt{Features} are assigned to a \texttt{Class}; the model with the highest CRA index value is found. The problem has an additional constraint requiring that each Feature is assigned to only one \texttt{Class} at a time.

\begin{wraptable}{R}{4.9cm}
\caption{Summary of CRA input models}
	\setlength{\tabcolsep}{0.3em}
	\scalebox{0.85}{
\begin{tabular}{l|c c c c c}

    	\hline
		             & \textbf{A} & \textbf{B} & \textbf{C} & \textbf{D} & \textbf{E} \\
		\hline
		Attributes      & 5 & 10 & 20 &  40 &  80 \\
		Methods         & 4 &  8 & 15 &  40 &  80 \\
		Data Dep.       & 8 & 15 & 50 & 150 & 300 \\
		Functional Dep. & 6 & 15 & 50 & 150 & 300 \\
		\hline
	\end{tabular}%
	}
		\label{tab:cra-input-models}%
\end{wraptable} 

The CRA case study authors provide a set of five input models. The difference between them is the number of Features present. Model A, is the smallest model with only nine features. The largest model provided is model E with 160 features. Across all models, each set of features has an increasing number of dependencies between them. A summary of all the input models is included in Table~\ref{tab:cra-input-models}. 

We are specifying the CRA case study using two sets of transformations. The first set is implemented manually, and consists of four operators as suggested in \cite{BurduselZschaler16}. Other TTC'16 participants that used a similar approach to solve the case studies used similar rules \cite{nagyclass,Fleck+16_TTC_Case}. The second set of operators are aCPSOs generated using the approach presented in this paper. 

\subsubsection{Scrum Planning}

	We are running two experiments for the Scrum Planning (SP) case study described in Sect. \ref{section:running_example}. This case study has a similar problem specification as the CRA case study with the following differences: the assigned items do not have any dependencies between each other as \texttt{Features} do in the CRA case, and this case study is specified as a multi-objective problem.

	In Table~\ref{tab:input-models-summary} we include a description of the input models used in experiments for this case study. These have been automatically generated by the authors using a random model generator. 
	Through this case study evaluation, we explore how the difference in the number of objective functions affects the behaviour of manual and generated rules.

\subsubsection{Next Release Problem}

	The aim of the Next Release Problem (NRP) is to find the optimal set of tasks to include in the next release for a software product, to minimise the cost and to maximise the customer satisfaction~\cite{Zhang2007MOnextReleaseProblem}. Each \texttt{Customer} has a desire which can consist of one or many \texttt{SoftwareArtifacts}. \texttt{SoftwareArtifacts} can have a recursive dependency on other \texttt{SoftwareArtifacts}.

\begin{table}[!t]
\centering
\caption{Summary of input models for SP and NRP case studies.}
\label{tab:input-models-summary}
\begin{tabular}{|l|l|l||l|l|l|}
\hline
\multicolumn{3}{|c||}{\textbf{Scrum Planning}} & \multicolumn{3}{c|}{\textbf{Next Release Problem}} \\ \hline
Input Model      & A       & B       & Input Model            & A      & B       \\ \hline
Stakeholders     & 5       & 10      & Customers              & 5      & 25      \\ \hline
WorkItems        & 119     & 254     & Requirements           & 25     & 50      \\ \hline
Backlog Size     & 455     & 1021    & Software Artifacts     & 63     & 203     \\ \hline
\end{tabular}
\end{table}

To solve this problem, the user is required to assign instances of \texttt{SoftwareArtifacts} to a \texttt{NextRelease} such that the total cost of the selected \texttt{SoftwareArtifacts} is minimised and the total customer satisfaction is maximised. We are specifying the next release problem using two sets of evolvers. One set was manually designed by the third author, who was not involved in developing the rule-generation algorithm. 
The second set uses the automatically generated CPSOs, using the approach described in this paper.

The minimal set of required rules for this case study is simple, only requiring mutations to add and remove an edge between a \texttt{Solution} and a \texttt{SoftwareArtifact}. However, the difference between this case study and the others considered in this paper is that the selection of a \texttt{SoftwareArtifact}, directly influences the Cost fitness function and indirectly influences the Customer Satisfaction objective. A \texttt{SoftwareArtifact} is considered for the calculation of a \texttt{RequirementRealization}, only when all its dependencies are also assigned to the solution. With this difference, we aim to evaluate how the generated rules explore the search space in cases where the fitness functions provide only coarse--granular guidance.

The input models used for this case study have also been automatically generated by the authors using a random model generator. A brief description of these models has been included in Table~\ref{tab:input-models-summary}. 


\subsection{Experiment configurations}
\label{subsection:experiments}

We run experiments and compare the quality of the solutions obtained using two configurations: \man with manually specified mutation operators, and \gen with automatically generated mutation operators.
For multi--objective problems we use the hypervolume indicator and the ratio of best solutions for our comparison. For the CRA case, which is single--objective, we compare the quality of the solutions based on the median CRA score.
%

\paragraph{Experimental Setup}

	All the experiments have been repeated 30 times for statistical significance \cite{Arcuri2014Hitchhikers} and have been executed on Amazon Web Services (AWS) c5.large spot instances running Amazon Linux 2 build 4.14.101-75.76.amzn1.x86\_64 running Java version 11.0.3+7-LTS.
	We ran our experiments using the NSGA-II algorithm \cite{Deb2002}. The NSGA-II algorithm is a well established algorithm that has been used successfully in many SBSE applications \cite{Boussaid2017}. 

	We undertook our experiments in two stages. The first stage was for determining ideal algorithm parameters (hyperparameters) that worked well for both configurations.
	The second stage, used the hyperparameters found in the first stage, and was for comparing the quality of \gen with \man.

\paragraph{Hyperparameter Selection}
	
	 \begin{figure}[t]
	 	\centering
		
	 	\includegraphics[width=\linewidth]{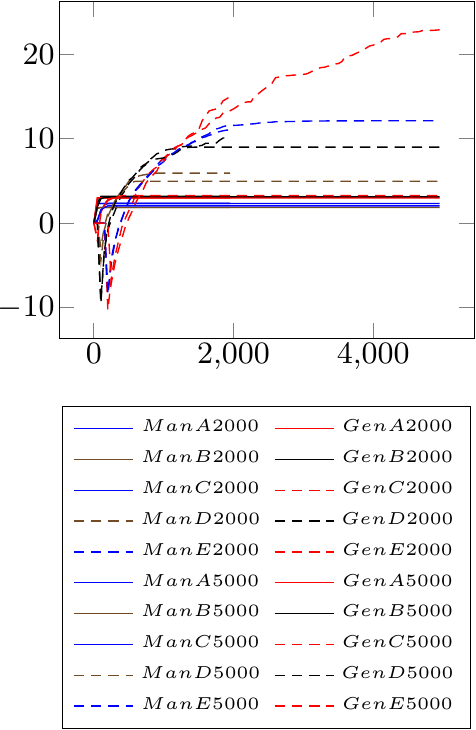}%
	 	
	 	\caption{Parameter search runs for the CRA case study. The X axis shows the number of fitness evolutions, and the Y axis shows the median objective calculated across all batches.}
	 	\label{fig:evolutions_parameters_cra}
	 \end{figure}

	 \begin{figure}[t]
		\centering

		\includegraphics[width=\linewidth]{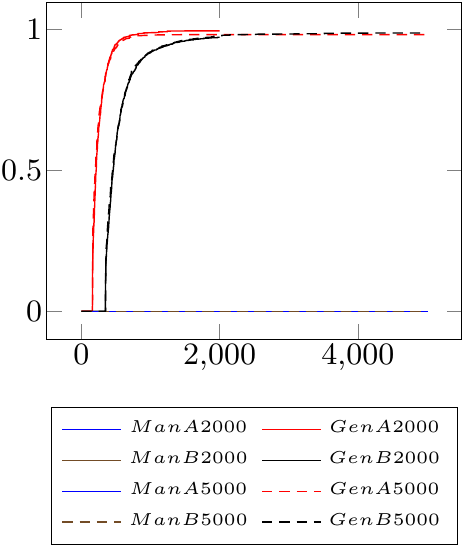}%
		
		\caption{Parameter search runs for the SP case study. The X axis shows the number of fitness evolutions, and the Y axis shows the median hypervolume.}
		\label{fig:evolutions_parameters_sp}
	\end{figure}

	\begin{figure}[t]
		\centering
	   
		\includegraphics[width=\linewidth]{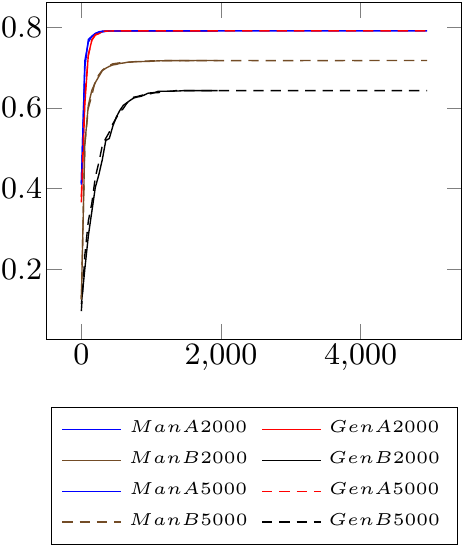}%

		\caption{Parameter search runs for the NRP case study. The X axis shows the number of fitness evolutions, and the Y axis shows the median hypervolume.}
		\label{fig:evolutions_parameters_nrp}
	\end{figure}

	We \change{systematically searched}{performed a systematic search} for ideal population\--size and number\--of\--evolutions hyperparameters that allow each configuration to find the best solutions. The combinations of analysed parameter configurations have each been repeated 10 times to ensure robust results.

	To identify a good number of evolutions to use in our experiments, we set the population size to 100 solutions, and analysed the growth of the median objective value for the single--objective problems and median hypervolume for the multi--objective problems and selected the number of evolutions after which there was no significant increase until the number of fitness evaluations has been exhausted and the algorithm stopped. After we have selected the number of evolutions based on the plateau of the fitness functions, we tried to reduce the size of the population by applying decrements of 25, until we reached a population size of 50. However, upon evaluating the results across all case studies, we determined the population size of 100 to be the best value for our experiments. 

	In Fig.~\ref{fig:evolutions_parameters_cra} we show the evolution of the median CRA objective found by configurations with 2000 and 5000 evolutions respectively and population size 100. We observe that for all input models, except for E, the CRA index value plateaus after 2000 evolutions. For input model E, the \gen configuration continues to increase, even after the \man configuration starts to plateau after passing 2000 evolutions. 
	
	Fig.~\ref{fig:evolutions_parameters_sp} shows a summary of the hyperparameter runs for the SP case study. We observe that \man is getting stuck in more than half of the experiment repetitions and this leads in a median HV of 0 for this configuration. The \gen configuration is consistent at finding good solutions with a high HV value, and starts to plateau after 2000 evolutions.

	In Fig.~\ref{fig:evolutions_parameters_nrp} we show the evolution of the median HV by configurations with 2000 and 5000 evolutions and population size 100 for the NRP case. We observe that for input model A, there is no noticeable difference between \man and \gen. For input model B, \man finds a higher HV metric than \gen. We also observe that all configurations stop finding solutions after 500 evolutions for model A and 1000 evolutions for model B.

	Based on the results of the experiment configurations discussed in this section we have selected the algorithm parameters for the experiment configurations used in our experiments. The selected number\--of\--evolutions parameter values have been included in the Evol column in Table~\ref{tab:cra_results} for the CRA case, Table~\ref{tab:sp_mo_results} for the SP case and Table~\ref{tab:nrp_mo_results} for the NRP case.

\paragraph{Hypervolume indicator}

	Comparing solutions of optimisation problems that have more than one objective value is not a trivial problem. This is because when one objective value changes, the value of the other objectives can change as well. To overcome this problem, the hypervolume unary indicator has been proposed in~\cite{zitzler1998hypervolume}.
	This single--value metric, measures the dominated volume between the solution points belonging to the Pareto front and a reference point (also nadir point) defined by the objective values of the worst solution. Higher hypervolumes indicate a Pareto front closer to the theoretical optimum.

\paragraph{Ratio of Best Solutions}

	For multi--objective problems we are calculating the \emph{Best Solutions Ratio} (BSR) to show the number of non-dominated solutions contributed to the Pareto front by each configuration as presented by \cite{Hansen1998}. In our approach we are building the reference set (RS) using the best solutions found by all runs for both configurations. This metric allows us to measure the percentage of the contributions made to the reference set by each configuration.

	\begin{equation}
		BSR = {{|S \cap PF_{pseudo}|} \over {|PF_{pseudo}|}}
	\end{equation}

	$PF_{pseudo}$ stands for the reference set obtained by merging all the known nondominated solutions for a problem instance. $S$ stands for the reference set of the configuration for which the metric is being calculated.

\paragraph{Statistical Analysis}

	We use the Mann-Whitney U test to perform a statistical analysis of our results~\cite{mann1947statistical}. To measure the size of the differences between the configurations we use Cohen's d effect metric \cite{cohen1988statistical}. We are also including standard deviation (SD), skewness (Skew) and Kurtosis (Kurt) in our results tables to give a better indication of the solutions distribution found in our experiments.

	 \section{Results}
\label{section:results}

In this section we present our experiment results for each of the case studies introduced in the previous section. We discuss each case study individually below. \add{The complete data set discussed in this section can be downloaded from \cite{dataset}.}

\subsection{Class Responsibility Assignment}
\label{section:class_responsability_assignment}
\begin{table}[!t]
\centering
\caption{Summary of CRA mutation operators for \man and \gen.}
\label{tab:cra-man-gen-operators}
\scalebox{0.85}{
\begin{tabular}{|l|l|}
\hline
\textbf{Manual}    & \textbf{Gen aCPSO}     \\ \hline
Create Class       & Create Class           \\ \hline
N/A                & Create Class Lb Repair \\ \hline
Assign Feature     & Assign Feature     	\\ \hline
Change Feature     & Change Feature     \\ \hline
N/A				   & Remove Feature		 \\ \hline
Delete Empty Class & Delete Class                    \\ \hline
N/A                & Delete Class Lb Repair \\ \hline
\end{tabular}
}
\end{table}

\begin{table}[!t]
	\centering
\caption{CRA results for \man and \gen.}
 \label{tab:cra_results}
	\scalebox{0.85}{
		\begin{tabular}{|l|l|l|l|l|l|l|l|}
			\hline
			\textbf{Config} & \textbf{Evol} & \textbf{Median} & \textbf{Min} & \textbf{Max} & \textbf{SD} & \textbf{Skew} & \textbf{Kurt} \\ \hline
			\textbf{Man A}  & 500           & 2.333           & 0.850        & 3.000        & 0.552         & -0.679        & -0.509        \\ \hline
			\textbf{Gen A}  & 500           & \textbf{3.000}  & 3.000        & 3.000        & 0.000         & 0             & 0             \\ \hline \hline 
			\textbf{Man B}  & 500           & 1.865           & 1.238        & 3.104        & 0.514         & 0.642         & -0.032        \\ \hline
			\textbf{Gen B}  & 500           & \textbf{3.167}  & 1.826        & 4.083        & 0.599         & -0.470        & -0.376        \\ \hline \hline
			\textbf{Man C}  & 500           & 2.224           & 1.148        & 3.240        & 0.572         & -0.089        & 0.824         \\ \hline
			\textbf{Gen C}  & 500           & \textbf{3.129}  & 2.110        & 3.806        & 0.428         & -0.539        & -0.039        \\ \hline \hline
			\textbf{Man D}  & 2000          & 5.191           & 3.557        & 7.041        & 0.837         & 0.068         & 0.339         \\ \hline
			\textbf{Gen D}  & 2000          & \textbf{9.863}  & 7.634        & 12.273       & 1.257         & -0.176        & 0.782         \\ \hline \hline
			\textbf{Man E}  & 2500          & 11.572          & 8.879        & 14.691       & 1.639         & 0.122         & 0.663         \\ \hline
			\textbf{Gen E}  & 2500          & \textbf{17.323} & 11.698       & 20.051       & 1.604         & -1.106        & -3.176        \\ \hline
			\end{tabular}
	}
\end{table}

In Table~\ref{tab:cra-man-gen-operators} we list the mutation operators used for the two experiment configurations. 
	Both configurations use three identical mutation operators to create a class (Create Class) and assign and change a feature (Assign Feature, Change Feature).
		
	For \gen  the change \texttt{Feature} operator contains a PAC requiring that the source \texttt{Class} still has at least one \texttt{Feature} assigned following the application of this operator. At the same time, the \man operator to change a \texttt{Feature} can generate an empty class upon its application, and such instances are fixed by the delete empty class operator.
	In addition to these operators, \gen contains two additional operators to create and delete a \texttt{Class} after all the \texttt{Features} have been assigned. These ensure that the search does not get stuck in local optima in cases where the \texttt{Features} are assigned to too many or too few \texttt{Classes}. 
	%
	%
	The performance of the \gen operators is not affected.

	\begin{table}[!t]
		\centering
		\caption{Summary of statistical testing results for CRA.}
		\label{tab:cra_statistical_testing_results}
		\scalebox{0.85}{
		\begin{tabular}{|l|l|l|l|l|l|}
		\hline
		\textbf{} & \textbf{A} & \textbf{B} & \textbf{C} & \textbf{D} & \textbf{E} \\ \hline
		\textbf{p}-value & \textless{}0.05\% & \textless 0.05\% & 0.05\% & 0.05\% & 0.05\% \\ \hline
		\textbf{U}-value & 795 & 809.5 & 817 & 900 & 884 \\ \hline
		Cohen's \textbf{d} & Large & Large & Large & Large & Large \\ \hline
		\end{tabular}
		}
		\end{table}

		\begin{table*}[!t]
			\centering
			\caption{Summary of CRA elapsed time in seconds for the \man and \gen configurations across all input models.}
			\label{tab:cra-elapsed-time}
			\begin{tabular}{|l|l|l||l|l||l|l||l|l||l|l|}
			\hline
			\textbf{Time} & \textbf{Man A} & \textbf{Gen A} & \textbf{Man B} & \textbf{Gen B} & \textbf{Man C} & \textbf{Gen C} & \textbf{Man D} & \textbf{Gen D} & \textbf{Man E} & \textbf{Gen E} \\ \hline
			\textbf{Mean} & 15.10 & 27.90 & 23.27 & 43.76 & 41.32 & 75.28 & 611.40 & 1177.70 & 2972.65 & 4298.16 \\ \hline
			\textbf{Median} & 14.92 & 27.61 & 22.04 & 44.24 & 41.07 & 75.65 & 590.75 & 1188.91 & 2869.16 & 4198.20 \\ \hline
			\textbf{Min} & 11.44 & 26.48 & 17.33 & 33.92 & 26.79 & 64.19 & 452.90 & 991.29 & 2193.35 & 3582.67 \\ \hline
			\textbf{Max} & 17.64 & 30.09 & 34.99 & 50.13 & 58.43 & 86.35 & 853.47 & 1416.96 & 4202.11 & 5189.39 \\ \hline
			\end{tabular}
		\end{table*}

  	Table~\ref{tab:cra_results} shows summary statistics for the CRA index found using the two configurations.
	For all input models, the configuration with automatically generated rules (\gen) consistently finds better median CRA index values than the configuration with manual rules (\man). In all cases, \gen also finds higher minimum (Min) and maximum (Max) CRA scores than \man. These results are confirmed by Table~\ref{tab:cra_statistical_testing_results} which shows the \emph{p} and \emph{U} values of the Mann-Whitney test and Cohen's \emph{d} effect size.
	
	The quality of the results found by \gen is attributed to the aCPSO operators which allow classes to be created and deleted, after all the features have been assigned to a class, without invalidating the multiplicity constraints.
	The results for this experiment show that our approach is good at generating mutation operators that help find results that are comparable to manually specified mutation operators.
	%

\begin{figure}[!b]
	\centering
	\subcaptionbox{Input Model A\label{fig:pareto_front_sp_a}}{\includegraphics[width=0.48\linewidth]{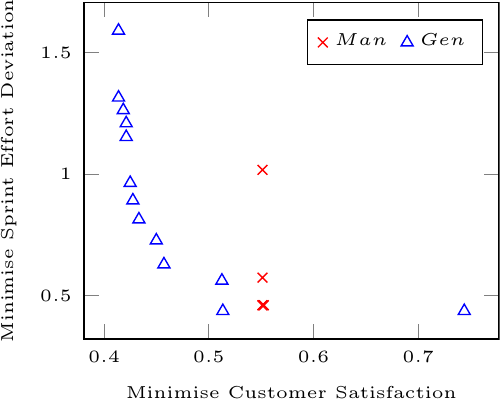}}%
	\hspace*{\fill}
	\subcaptionbox{Input Model B\label{fig:pareto_front_sp_b}}{\includegraphics[width=0.48\linewidth]{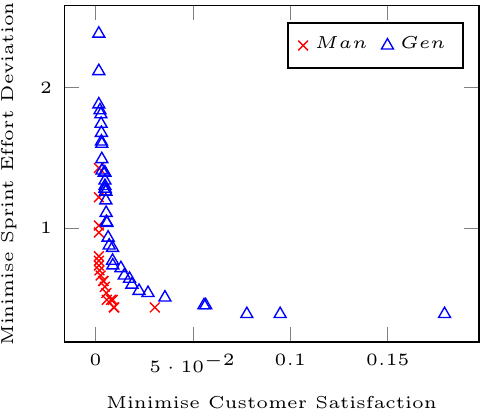}}%
	\hspace*{\fill}
	\caption{Scrum Planning Pareto fronts}
	\label{fig:pareto_fronts}
\end{figure}

\subsection{Scrum Planning}
\label{section:scrum_planning}

	\begin{table}[!t]
		\centering
        \caption{SP results for \man and \gen configurations.}
        \label{tab:sp_mo_results}
        \scalebox{0.85}{
        \begin{tabular}{|l|l|l|l|l|l|l|l|l|}
        \hline
        \textbf{Config} & \textbf{Evol} & \textbf{Median} & \textbf{Min} & \textbf{Max} & \textbf{SD} & \textbf{RS} & \textbf{RSC} & \textbf{BSR} \\ \hline
        \textbf{Man A}  & 1500          & 0.000           & 0.000        & 0.960        & 0.460       & 13        & 0           & 0.00        		\\ \hline
        \textbf{Gen A}  & 1500          & \textbf{0.959}  & 0.957        & 0.995        & 0.010       & 13        & 13          & \textbf{1.00}        		\\ \hline \hline
        \textbf{Man B}  & 2500          & 0.492           & 0.000        & 0.996        & 0.505       & 25        & 19          & \textbf{0.76}        \\ \hline
        \textbf{Gen B}  & 2500          & \textbf{0.988}  & 0.983        & 0.998        & 0.004       & 25        & 6           & 0.24        		\\ \hline
        \end{tabular}
        }
		\end{table}
		
		\begin{table}[!t]
			\centering
			\caption{Summary of SP elapsed time in seconds for the \man and \gen configurations across all input models.}
			\label{tab:sp-elapsed-time}
			\begin{tabular}{|l|l|l|l|l|}
			\hline
			\textbf{Time} & \textbf{Man A} & \textbf{Gen A} & \textbf{Man B} & \textbf{Gen B} \\ \hline
			\textbf{Mean} & 119.13 & 291.25 & 484.90 & 3069.18 \\ \hline
			\textbf{Median} & 120.67 & 291.87 & 487.76 & 3016.74 \\ \hline
			\textbf{Min} & 107.63 & 261.25 & 447.16 & 2686.92 \\ \hline
			\textbf{Max} & 130.47 & 367.78 & 510.77 & 4171.07 \\ \hline
			\end{tabular}
			\end{table}

\begin{table}[!t]
\centering
\caption{Summary of SP mutation operators for \man and \gen.}
\label{tab:sp-man-gen-operators}
	\scalebox{0.85}{
		\begin{tabular}{|l|l|}
			\hline
				\textbf{Manual}     & \textbf{Gen aCPSO}        \\ \hline
				Create Sprint       & Create Sprint             \\ \hline
				N/A                 & Create Sprint Lb Repair   \\ \hline
				Add WorkItem        & Add WorkItem              \\ \hline
				Change WorkItem     & Change WorkItem           \\ \hline
					N/A				& Remove WorkItem		  	\\ \hline
				Delete Empty Sprint & Delete Sprint             \\ \hline
				N/A                 & Delete Sprint Lb Repair   \\ \hline
		\end{tabular}
	}
\end{table}

	The SP case study is specified as a multi-objective problem. To compare the results we will use the hypervolume metric.
	In Table~\ref{tab:sp-man-gen-operators} we include the mutation operators used for the two experiment configurations. These mutation operators are similar to the ones generated for the CRA case study.
	
  	Table.~\ref{tab:sp_mo_results} shows a comparison of the calculated hypervolumes for this case study for input models A and B.
	For both input models \man finds fewer constraint satisfying solutions, compared to \gen. For model A, \man only found valid solutions in 10 out of the 30 experiment runs, compared to \gen  which found no invalid solutions. For cases where only invalid solutions have been found, we allocated a value of 0 for the hypervolume, because there are no constraint satisfying solutions generated and the covered hypervolume space in those cases is 0.
	The same effect can be observed for model B, for which \man finds valid solutions for 15 out of 30 experiments, compared to \gen which always found a valid solution. Comparing the median hypervolume values for the configurations with valid solutions, we observe that \gen is better than \man for both input models.

	For model A, the highest hypervolume values have been found by \gen and all the reference set contributions (RSC) are generated by this configuration, which has a BSR rate of 100\%. In Fig.~\ref{fig:pareto_front_sp_a} we include the Pareto fronts found by the two configurations for model A, and in this figure we can see that that \gen's reference set solutions dominates all the solutions found by \man. The \man reference set contains 5 solutions, while the \gen one has 13. The Mann-Whitney U test shows that \gen is better than \man for model A (\emph{p}=4.266E-6, \emph{U}=761, Cohen's d=Large).
	For model B, \gen also found a higher median hypervolume value. However, for this model, \man found 19 out of 25 reference set solutions, giving it a BSR rate of 76\% while \gen only found 6 reference set solutions, with a BSR rate of 24\%. The Mann-Whitney U test shows that \gen is as good as \man for model B (\emph{p}=0.25, \emph{U}=527, Cohen's d=Large).
	In Fig.~\ref{fig:pareto_front_sp_b} we include the Pareto fronts found by \gen and \man for model B. As indicated by the BSR rate, \man finds more dominating solutions than \gen in the runs that found valid solutions, however \gen found are more diverse solutions that cover a wider area along the Pareto curve. The \man reference set contains 19 solutions, while the \gen one has 40 solutions. 
	
	We believe that \man is getting stuck in local optima and the operators are unable to explore new solutions without temporarily invalidating or decreasing the quality of the current solutions. At the same time \gen is able to explore new solutions without invalidating the constraints, but it is also affected by being stuck in local optima. \add{We attribute the better results for model B found by \man to the fact that after constraint satisfying solutions are discovered, the best solutions are found by moving \emph{WorkItem} elements between \emph{Sprints}, until the right configuration is found.}

	For this case study, \gen finds a consistently good hypervolume, with a small SD value across all the repetitions, as seen in the Median and SD columns in Table.~\ref{tab:sp_mo_results}. The difference between the numbers of valid solutions found, shows that the addition of the aCPSOs helped the search to find consistent solutions.

\subsection{Next Release Problem}
\label{section:next_release_problem}

	\begin{table}[!t]
		\centering
		\caption{NRP results for \man and \gen.}
		\label{tab:nrp_mo_results}
		\scalebox{0.85}{
		\begin{tabular}{|l|l|l|l|l|l|l|l|l|}
		\hline
		\textbf{Config} & \textbf{Evol} & \textbf{Median} & \textbf{Min} & \textbf{Max} & \textbf{SD} & \textbf{RS} & \textbf{RSC} & \textbf{BSR} \\ \hline
		\textbf{Man A}  & 750           & 0.791           & 0.791        & 0.791        & 0.000       & 32        & 32          & 1.00        \\ \hline
		\textbf{Gen A}  & 750           & 0.791           & 0.791        & 0.791        & 0.000       & 32        & 32          & 1.00        \\ \hline \hline
		\textbf{Man B}  & 1500          & \textbf{0.718}  & 0.712        & 0.722        & 0.003       & 281       & 281         & \textbf{1.00}        \\ \hline
		\textbf{Gen B}  & 1500          & 0.641           & 0.635        & 0.643        & 0.002       & 281       & 63          & 0.22        \\ \hline
		\end{tabular}
		}
	\end{table}

	\begin{table}[!t]
		\centering
		\caption{Summary of NRP elapsed time in seconds for the \man and \gen configurations across all input models.}
		\label{tab:nrp-elapsed-time}
		\begin{tabular}{|l|l|l|l|l|}
		\hline
		\textbf{Time} & \textbf{Man A} & \textbf{Gen A} & \textbf{Man B} & \textbf{Gen B} \\ \hline
		\textbf{Mean} & 275.42 & 223.42 & 1677.80 & 1355.29 \\ \hline
		\textbf{Median} & 274.96 & 224.27 & 1676.22 & 1348.97 \\ \hline
		\textbf{Min} & 258.84 & 215.79 & 1610.85 & 1312.45 \\ \hline
		\textbf{Max} & 307.71 & 234.52 & 1813.63 & 1412.93 \\ \hline
		\end{tabular}
		\end{table}

	\begin{table}[!t]
		\centering
		\caption{Summary of NRP mutation operators for \man and \gen.}
		\label{tab:nrp-man-gen-operators}
		\scalebox{0.85}{
			\begin{tabular}{|l|l|}
			\hline
			\textbf{Manual}             & \textbf{Gen aCPSO}           \\ \hline
			Modify SA With Dependencies & N/A                      \\ \hline
			Modify Software Artifact    & N/A                      \\ \hline
			Assign Highest Realisation  & N/A                      \\ \hline
			Fix Dependencies			& N/A					\\ \hline
			N/A                         & Add Software Artifact    \\ \hline
			N/A                         & Remove Software Artifact (PAC) \\ \hline
			\end{tabular}
		}
	\end{table}

\begin{figure}[!b]
	\centering
	\subcaptionbox{Input Model A\label{fig:pareto_front_nrp_a}}{\includegraphics[width=0.48\linewidth]{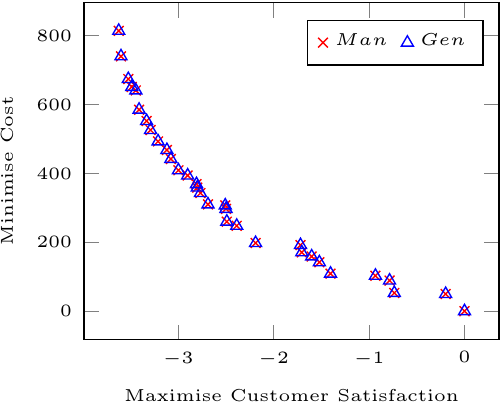}}%
	\hspace*{\fill}
	\subcaptionbox{Input Model B\label{fig:pareto_front_nrp_b}}{\includegraphics[width=0.48\linewidth]{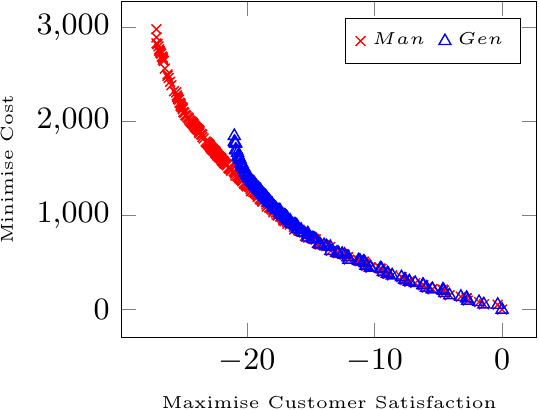}}%
	\hspace*{\fill}
	\caption{Next Release Problem Pareto fronts.}
	\label{fig:nrp_pareto_fronts}
\end{figure}

	In contrast to the other use cases, the operators used in the NRP case are substantially different for both configurations (Table~\ref{tab:nrp-man-gen-operators}). The \gen operators only cover the basics: addition and removal of single \texttt{SoftwareArtifacts}. In both situations, chances are that costs are raised without improving customer satisfaction due to the introduction of missing dependencies.

	The first operator of \man overcomes this problem, only adding a \texttt{SoftwareArtifact} if all of its dependencies are already part of the solution. Likewise, the removal of an artifact is only possible if no dependent artifacts are left over. The second operator allows for larger steps through the search space by adding and removing \texttt{SoftwareArtifacts} together with their direct dependencies and dependent artifacts, respectively. \texttt{Assign Highest Realisation} tries to exploit the fact that among \texttt{Realisations} of the same \texttt{Requirement} those with the highest \texttt{percentage} contribute most to the customer satisfaction. Considering all \texttt{Realisations} with yet unfulfilled dependencies, the operator selects the one with the highest \texttt{percentage} of fulfillment and adds its missing \texttt{SoftwareArtifacts} to the solution. 

	Note that none of the aforementioned operators take transitive dependency relations into account. Therefore, to counter the emergence of missing dependencies, the last operator is responsible for either adding all dependencies of an already selected \texttt{SoftwareArtifact} or removing the dependent artifacts of a formerly removed \texttt{SoftwareArtifact}.

	%
	For model A, we can observe that \gen consistently finds a hypervolume that is identical to to \man.
	Both configurations find the same solutions forming the reference set and each has a BSR rate of 100\%. We can also observe that the standard deviation metric between the hypervolumes across all 30 runs for each configuration is 0. This can also be observed in Fig.~\ref{fig:pareto_front_nrp_a} which includes the identical Pareto fronts for both configurations. Because the solutions found are identical for this model we are not including the statistical testing results in the paper, however these can be found in the data attachment.
	For model B, the hypervolume value found by \man is higher than the one for \gen. However, on a closer inspection of the generated Pareto fronts for both configurations in Fig.~\ref{fig:pareto_front_nrp_b}, we see that the solutions found by \gen, are subsets of the fronts for the \man configuration. This is also confirmed by the data in Table~\ref{tab:nrp_mo_results}. \man has a BSR rate of 100\%, while \gen only has a BSR rate of 22.4\%. In this case \man also includes all the solutions generated by \gen. For model B the Mann-Whitney test shows at 5\% confidence level, that \man finds solutions of better quality than \gen, with a large effect size.

	We attribute this behaviour to the way in which the \man operators have been designed, compared to the ones automatically generated. The \man operators, ensure that a \texttt{SoftwareArtifact}, together with all its dependencies are all assigned to a Solution in a single application. At the same time, the \gen operators, assign \texttt{SoftwareArtifacts}, one by one, by adding or removing edges, in atomic operations. Because the Customer Satisfaction objective does not provide guidance for the solutions, unless all \texttt{SoftwareArtifacts} realising a complete Realization are assigned, \gen is slower at finding converging solutions, requiring more evolutions. However comparing the structure of the operators, a single operation of the manual operator which moves a \texttt{SoftwareArtifact} together with all its dependencies in a single evolution, is equivalent to running the \gen configuration operator to add an edge, for multiple evolutions. We are currently investigating how different strategies for adjusting the step size can help aCPSOs overcome this issue and the initial results show that with the correct step size selection strategy the search algorithm can overcome limitations similar to the one observed in this case.
	
	\subsection{Performance}

	\label{section:performance}
		
		We have also compared the efficiency of the generated operators with the manually created ones. In all cases but one, the generated operators led to shorter (or at most equal) average runtimes for the ES. For two scenarios the generated rules were less efficient than the manual ones. This happened for the CRA and SP case studies.

		\add{In Table~\ref{tab:cra-elapsed-time} we include a runtime summary for the two configurations we are evaluating across all input models for the CRA case study. We observe a higher runtime for \gen configurations, that is almost double the time required by the \man configuration. We attribute this difference to the higher number of rules used by the \gen configuration.} For CRA, we additionally explored two different matching strategies in MDEOptimiser: the \emph{classic strategy}\footnote{Which the tool authors used in their submission to TTC'16~\cite{BurduselZschaler16}} first finds all possible matches for all operators and then uniformly randomly selects one of them, while the \emph{non-deterministic matching strategy} uses Henshin's non-deterministic matching algorithm by uniformly randomly selecting one mutation operator and then letting Henshin apply this for a random match. For the CRA case, there are more generated operators than manual ones, which means that more matches must be generated in the `classic strategy'. As a result, under this strategy the search with generated rules took up to approximately 3 times as long as with manual rules. With the `non-deterministic matching strategy', the generated rules led to a faster search than the manual rules.
		
		\add{In Table~\ref{tab:sp-elapsed-time} we include the a summary of the runtime for the SP case study. We observe that \gen is slower than \man. After closely inspecting the generated results we observed that \gen finds more constraint satisfying solutions and more time is spent evaluating the fitness functions and at the same time the NSGA-II archive contains more solutions for the \gen configuration compared to \man, which results in more time being required to perform the required domination and crowding comparisons. This leads to an increase in the runtime for \gen.}

		\delete{The runtimes for both configurations for all case studies have been included in the data attachment accompanying the paper.}

	\subsection{Threats to validity}

		The validity of the conclusions we draw from our data depends on a number of factors: 
			\begin{enumerate*}
				\item the degree to which the selected case studies are representative of real-world problems, 
				\item the chosen hyperparameters (e.g., population size, number of evolutions, ...), 
				\item the degree to which the chosen input models are representative of real-world problems, and 
				\item the provenance of the manual rules used in our experiments.
			\end{enumerate*}

		We have used a varied selection of case studies that cover both single and multi objective scenarios and allow a systematic exploration of different aspects of the overal problem. All hyperparameter values have been selected systematically to ensure that no approach is favoured over the other. We have applied the recommended steps to ensure that our results are accurate and correctly interpreted and described~\cite{Arcuri2014Hitchhikers}. Input models have been either provided as part of pre-existing case studies (CRA~\cite{Fleck+16_TTC_Case}), or have been randomly generated, ensuring consistency with the given problem metamodel. Recently, better model generators have been proposed~\cite{semerath2018graph, schneider2017graph} that aim to produce more realistic model instances for such evaluations. We are interested to explore the use of such generators for further evaluation of our approach. Finally, the manual rules that we use in our experiments were all produced without consideration of the generative principles we propose in this paper: the CRA rules were produced by the authors in 2016~\cite{BurduselZschaler16}, well before we started considering automatic generation of rules; the SP rules are very similar to the CRA rules. The NRP rules were produced by the 3rd author taking into account the structure of the objective functions during rule construction.

	\section{Related work}
\label{section:related_work}

	\paragraph{Mutation Generation}
	The generation of mutation operators for evolutionary algorithms has been studied in the wider optimisation literature. To the best of our knowledge, FitnessStudio~\cite{Struber2017} is the only approach in an MDE context. FitnessStudio is a meta-learning tool for generating in-place model transformation rules that can be used as search operators in model based optimisation. The algorithm generates mutation operators that obtain good results for the CRA case study~\cite{FleckTW16}. The main drawback is that the user is required to first execute a learning operation on a test model and then run the optimisation with the generated rules on the rest of the models that have to be optimised. The effectiveness of the approach depends on the model used for learning, its coverage of the metamodel and its similarity to the remaining models. In contrast to FitnessStudio, our approach does not require the additional meta-learning step.

	In \cite{HONG2018}, the authors present an offline hyper-heuristic approach that automatically generates mutation operators using genetic programming and meta-learning. These are then used in evolutionary programming with the aim of solving a number of test functions. This technique requires an already existing genetic encoding of the problem. In contrast, we support problems that are naturally encoded in a suitable DSML. The work in \cite{HONG2018} is similar to \cite{Struber2017}, requiring  a training step to first generate the mutation operators, which are then used to solve other problems. Our algorithm generates the mutation operators using the problem specification and does not require a training step.

	In \cite{alhwikem2016systematic} the authors introduce an approach for generating mutation operators for MDE languages. The goal of this approach is to use the generated operators to generate test inputs when performing mutation analysis. The systematically generated mutations can be used to change features of Ecore based languages, by adding, removing or changing values of a model feature, in order to increase test coverage. 
	The atomic mutations generated by this approach are similar to some operations we generate for the simple cases, namely to add, remove and change an element. In addition to these operators, our approach also generates more complex repairs.
	
	Mengerink et al. in~\cite{mengerink2016complete} propose a complete DSL operator library for EMF based languages. The operators are atomic operators and the proposed list of the most used operators also contains the aCPSO and CPSO operators generated by our approach and the SERGe rules generator. This library aims to be a complete list of all the possible atomic mutation operators for EMF based languages. The important contribution we make is to selectively generate only those operators useful in the context of ES.	
		
	\paragraph{Mutation Weighting}

	In \cite{doerr2016right}, the authors propose the use of operator strengths to increase the degree of changes performed by an operator. The authors show that a combination of atomic changes combined with variably sized changes, is the best for increasing the speed of solving optimisation problems. Using only atomic operators, the search can be slow, requiring many steps to be performed, while using operators that perform bigger changes, the search can have difficulty in finding neighboring solutions that have a better fitness. 
	Our case study evaluation showed that this is also a problem affecting our approach. For NRP and the case study presented in \cite{murphy2018deriving}, the generated aCPSOs require more applications to find good solutions, compared to operators that perform multiple operations in a single step. One potential solution to this problem is increasing the number of evolutions, and at the expense of longer runtime, the search can find better solutions if the fitness functions can efficiently guide each aCPSO application. Alternatively, the problem can be solved using a combination of operators consisting of aCPSOs and compositions of multiple aCPSOs that are applied in a single step. This is a problem we seek to explore in future work to expand our mutation generation approach.

	\section{Conclusions}
\label{section:conclusions}

	We have shown how mutation operators for search-based model engineering can be generated automatically without the need for meta-learning. The efficiency and effectiveness of the atomic consistency-preserving search operators (aCPSOs) we generate is comparable to search operators manually specified by expert users (and better in some cases). However, automatic generation requires less human effort and reduces the risk of accidentally introduced errors. 
	
	Our generated rules coped well with single- and multiple-objective problems as well as with a problem where the objective function provides only fairly coarse-grained guidance to the search. However, improvements are clearly possible. In particular, in our future work we plan to investigate the following questions:
	\begin{itemize}
		\item In the CRA case study, we have seen that the start-up behaviour of our generated rules differs from that of the manual rules, such that the manual rules find better solutions in early evolutions. We will study what affects this startup behaviour, and how we may be able to improve our generated rules in this area. For example, it may be useful to use separate sets of rules for the two phases of the search (cf. Sect.~\ref{section:proposed_method:requirements}) to ensure more focused exploration during the first phase.
		\item Optimisation problems use other constraints, beyond multiplicities. 
		Arbitrary constraints are difficult to handle without additional user input, however specific types of constraints or constraint templates can be more easily incorporated. For example, we are currently working on implementing rule generation for feature-model constraints.
		

		\item Recursive repair, offers additional opportunities for repair in CPSOs, but at the cost of higher generation effort and a larger set of search operators. Which, if any, recursive repair strategies offer benefits to the overall search?
	\end{itemize}

	\section*{Acknowledgment}
	This work has been supported by the \textit{Engineering and Physical Sciences Research Council} (EPSRC) under grant number 1805606.

    \bibliographystyle{ieeetr}
  	\bibliography{biblio}

\end{document}